\documentclass[10pt,journal,compsoc]{IEEEtran}
\ifCLASSOPTIONcompsoc
  \usepackage[nocompress]{cite}
\else
  \usepackage{cite}
\fi

\usepackage{booktabs}
\usepackage{threeparttable}  
\usepackage{diagbox}
\usepackage[font=bf]{caption}
\usepackage{subcaption}
\usepackage{float}
\usepackage{amsmath}
\usepackage{amssymb}
\usepackage{array}
\usepackage{threeparttable}
\usepackage{multirow}
\usepackage{graphicx}
\usepackage{epsfig}
\usepackage{endnotes}
\usepackage{algorithm}
\usepackage[noend]{algpseudocode}
\usepackage[lettersize]{lscape}
\usepackage{makecell}
\usepackage{url}

\usepackage[shortlabels]{enumitem}
    \setlist[enumerate, 1]{1.}
    \setlist{leftmargin=*}
\usepackage{tikz}
\usepackage{epstopdf}
\usepackage{adjustbox}
\usepackage{soul}  
\usepackage[T1]{fontenc}  

\usepackage{titlesec}
\titleformat{\subsubsection}{\normalfont\normalsize\itshape}{\thesubsubsection}{1em}{}
\titlespacing*{\subsubsection}{0pt}{3.25ex plus 1ex minus .2ex}{0ex plus .2ex}

\usepackage[hidelinks,breaklinks=true]{hyperref}

\newcommand{\rawsystemname}{Mondrian}
\newcommand{\systemname}{{\textsc{Mondrian}}}
\newcommand{\approachname}{Compressive Packed Inference}
\newcommand{\safearea}{safe area}
\newcommand{\Safearea}{Safe area}
\newcommand{\safeareas}{safe areas}
\newcommand{\SafeArea}{Safe Area}
\newcommand{\SafeAreas}{Safe Areas}

\newcommand{\Packedframe}{Canvas}
\newcommand{\packedframes}{canvases}
\newcommand{\packedinference}{packed inference}
\newcommand{\Packedinference}{Packed inference}

\newcommand{\componentOne}{ROI Extractor}
\newcommand{\componentTwo}{Hybrid ROI Scale Estimator}
\newcommand{\estimatorOne}{Proactive Scale Predictor}
\newcommand{\estimatorTwo}{Reactive Scale Tuner}
\newcommand{\componentThree}{Packed Canvas Generator}

\newcommand{\singlecol}[2]{\multirow{#1}{*}{\begin{tabular}[c]{@{}c@{}}#2\end{tabular}}}

\begin{document}

\title{\huge \rawsystemname: On-Device High-Performance Video Analytics\\with Compressive Packed Inference}

\author{Changmin~Jeon, Seonjun~Kim, Juheon~Yi, and
        Youngki~Lee
        \IEEEcompsocitemizethanks{
        \IEEEcompsocthanksitem 
        Changmin Jeon is with the Department of Computer Science and Engineering, Seoul National University, Seoul, Korea (e-mail: wisechang1@snu.ac.kr).
        \IEEEcompsocthanksitem 
        Seonjun Kim is with the Department of Computer Science and Engineering, Seoul National University, Seoul, Korea (e-mail: cyanide17@snu.ac.kr).
        \IEEEcompsocthanksitem
        Juheon Yi is with Nokia Bell Labs, Cambridge, UK (e-mail: juheon.yi@nokia.com).
        \IEEEcompsocthanksitem 
        Youngki Lee is with the Department of Computer Science and Engineering, Seoul National University, Seoul, Korea (e-mail: youngkilee@snu.ac.kr).
        \IEEEcompsocthanksitem 
        Youngki Lee is corresponding author of this paper.
}
}

\IEEEtitleabstractindextext{%
\begin{abstract}

In this paper, we present {\systemname}, an edge system that enables high-performance object detection on high-resolution video streams.
Many lightweight models and system optimization techniques have been proposed for resource-constrained devices, but they do not fully utilize the potential of the accelerators over dynamic, high-resolution videos.
To enable such capability, we devise a novel \textit{{\approachname}} to minimize per-pixel processing costs by selectively determining the necessary pixels to process and combining them to maximize processing parallelism.
In particular, our system quickly extracts ROIs and dynamically shrinks them, reflecting the effect of the fast-changing characteristics of objects and scenes.
It then intelligently combines such scaled ROIs into large canvases to maximize the utilization of inference accelerators such as GPU.
Evaluation across various datasets, models, and devices shows {\systemname} outperforms state-of-the-art baselines (e.g., input rescaling, ROI extractions, ROI extractions+batching) by 15.0-19.7\% higher accuracy, leading to $\times$6.65 higher throughput than frame-wise inference for processing various 1080p video streams.
The code is available at \url{https://github.com/snuhcs/mondrian}.

\end{abstract}

\begin{IEEEkeywords}
Mobile deep learning, object detection, video analytics
\end{IEEEkeywords}

}
\maketitle
\IEEEdisplaynontitleabstractindextext
\IEEEpeerreviewmaketitle

\section{Introduction}

\begin{figure}[t]
    \centering
    \includegraphics[width=0.95\columnwidth]{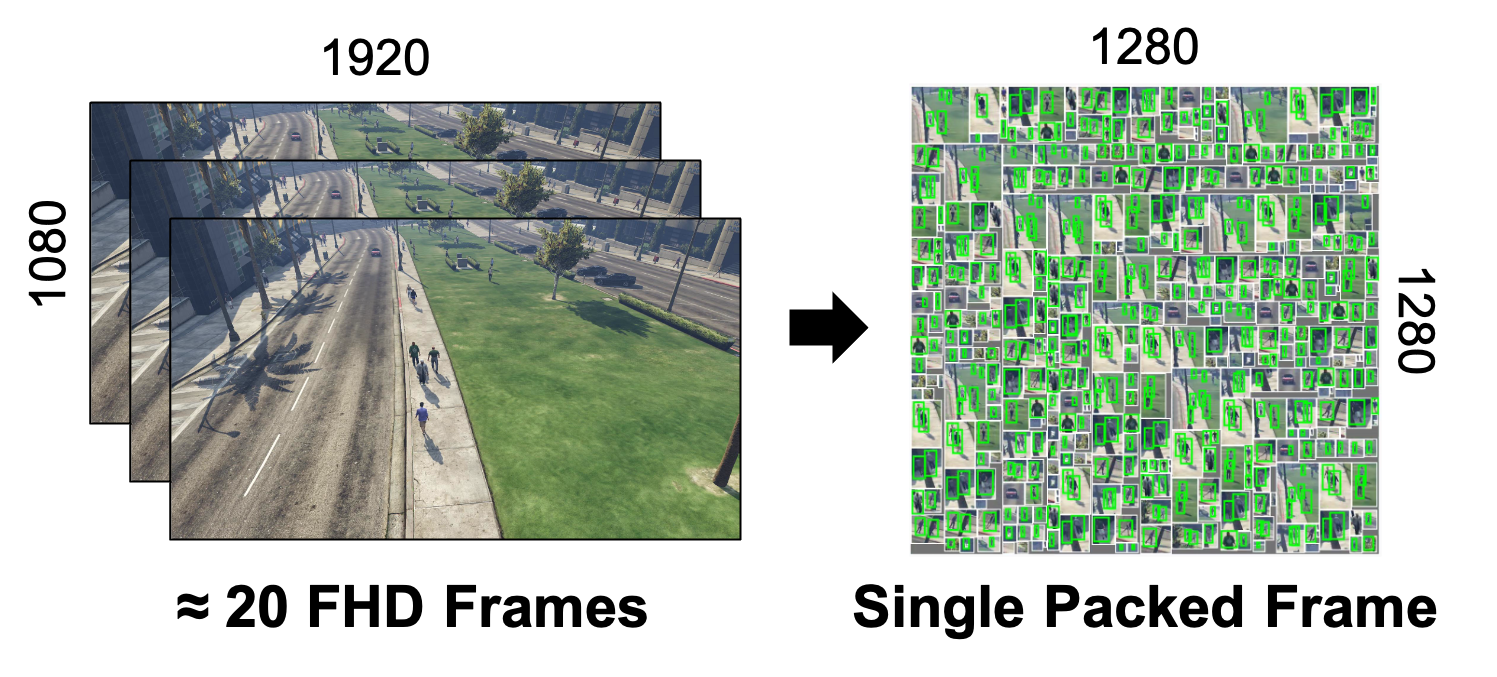}
    \caption{Concept of {\systemname}'s {\approachname}. We extract ROIs from 20 FHD frames, scale and pack them into a single 1280$\times$1280 canvas without accuracy drop.}
    \label{fig:intro-compressive-packed-inference}
\end{figure}

\IEEEPARstart{V}{ideo} analytics apps \cite{yi2020eagleeye, flexpatch} are quickly emerging with the advance of accurate Deep Neural Networks (DNNs) and high-performance accelerators such as GPU.
The edge-enabled systems have received increasing attention with their inherent advantages: (i) keeping private data in-house for camera operators to comply with privacy regulations (e.g., U.S. Reasonable Expectation of Privacy~\cite{privacy}) and (ii) reducing costs by avoiding network and cloud usage ($\approx$10 USD a day to process a 1080p video stream with a GPU~\cite{cloudPrice}).
However, the limited capacity of edge accelerators still constrains processing throughput over high-resolution video.
For example, a widely-used YOLOv5x~\cite{yolov5} object detector takes >1s ($\approx$0.53~FPS) on a high-end Adreno 730 GPU, and >250ms ($\approx$3.63~FPS) on a high-end Hexagon 780 DSP to process a 1080p frame.
The problems are aggravated as recent devices are equipped with multiple high-resolution cameras to cover a wide monitoring area ~\cite{natarajan2015multi} (Section~\ref{subsec:motiv-scenarios}).

It is a severe challenge to achieve high throughput (e.g., >30~FPS) for processing complex multiple high-resolution (e.g., 1080p) videos on-device.
There has been a stream of works for efficient object detection~\cite{apicharttrisorn2019frugal, zeng2020distream, he2021real}, but they achieve high performance only over simple videos with a limited number of objects and slow scene changes.
One common approach is to design lightweight DNNs (single-stage architecture~\cite{liu2016ssd, yolov5}, lightweight backbones~\cite{howard2017mobilenets}).
However, they cannot support high-resolution video processing on edge devices; for example, YOLOv5s achieves $\approx$30~FPS for 512$\times$512 video on Adreno 730 GPU, but the throughput quickly drops as resolution increases (Table~\ref{tb:lightweight-object-detectors}).
In another direction, a rich body of system optimization techniques accelerates high-resolution video processing through input resizing~\cite{chin2019adascale}, Region-of-Interest (ROI) extraction~\cite{liu2019edge, flexpatch, remix}, caching~\cite{huynh2017deepmon, xu2018deepcache}, and lightweight tracking~\cite{apicharttrisorn2019frugal, flexpatch}.
However, they continue to exhibit high \emph{per-pixel processing costs} and accuracy drop over dynamic videos (Section~\ref{subsec:motiv-prior-work-limitations}).
Recent papers~\cite{jeong2022band, seo2021slo} have proposed systems utilizing heterogeneous accelerators but without fully leveraging the unique characteristics of video analytics.


In this paper, we propose {\systemname}, a high-performance object detection system to process high-resolution videos with accelerators on the edge device.
Its core approach is \textit{\approachname} which minimizes \emph{per-pixel processing costs} by selectively determining the necessary pixels to process and combining them to maximize processing parallelism 
(Figure~\ref{fig:approach}).
The approach is enabled by two main sub-techniques.
First, it quickly extracts object-level ROIs and dynamically rescales them to \emph{\safeareas} (i.e., the smallest ROI without compromising detection accuracy).
Unlike prior ROI-based techniques, it dynamically rescales ROIs to reflect the fast-changing characteristics of the object and scene (e.g., object pose, motion, occlusion, lighting)
Second, with such scaled ROIs accumulated while SLO allows, it intelligently packs them into large frames (e.g., 1280$\times$1280), namely canvases, to maximize the utilization of mobile accelerators (e.g., GPU, NPU).


Designing {\systemname} involves the following challenges:

\vspace{3pt}
{\bf $\bullet$ High Variation of {\SafeAreas}.}
Our motivational study shows that extracted ROIs can be scaled to only 21\% of the original area on average (for the MTA dataset~\cite{mta}), but the {\safearea} of ROIs vary significantly (Figure~\ref{fig:approach-cdf-of-optimal-targets}), making the dynamic {\safearea} estimation a challenging problem. In particular, the {\safeareas} of ROIs exhibits significant spatial variations within a frame (objects with different sizes, poses, and levels of occlusions) and temporal variations across frames (object motions and lighting conditions).
With these compound factors, accurate yet lightweight estimation of per-ROI {\safearea} is a non-trivial problem.

\vspace{3pt}
{\bf $\bullet$ High Complexity in ROI Packing.}
Naively packing variable-sized ROIs results in a suboptimal combination of {\packedframes} with large wastage (unfilled pixels), significantly degrading the processing throughput.
However, finding an optimal combination is non-trivial, as this problem can be reduced to a 2D bin-packing problem~\cite{hartmanis1982computers} (NP-hard).
In addition, processing time and the total area of canvases exhibit non-linear relationships.
This further complicates the determination of the number and sizes of the canvases.
While some works~\cite{flexpatch, gokarn2023mosaic} have explored similar ideas, they do not consider the diverse application latency requirements, runtime inference latency fluctuation, and the effect of quantization which is critical to the system performance.
Hence, we need a careful problem formulation to find an efficient combination of {\packedframes}, packing layouts of the ROIs, and their assignments to accelerators.

We tackle these challenges with two key techniques:

\vspace{3pt}
{\bf $\bullet$ {\componentTwo} (Section~\ref{subsec:scale-estimator}).}
There are two possible approaches for {\safearea} estimation: (i) \emph{proactive prediction} (i.e., estimating the {\safearea} prior to inference using lightweight input features)~\cite{chin2019adascale, xu2020approxdet, xu2022litereconfig} and (ii) \emph{reactive probing} (i.e., running the object detector inference over multiple rescaled ROIs redundantly)~\cite{jiang2018chameleon, zhang2018awstream}.
Prediction is lightweight but hard to achieve high accuracy due to the difficulty of DNN accuracy prediction (works only for simple content at a coarse-grained scale~\cite{chin2019adascale}).
Probing is accurate but incurs high overhead to profile large search space (e.g., the rescale ratio ranges from 1$\times$ to 0.1$\times$ in the MTA dataset~\cite{mta}).
We take a novel hybrid approach to estimate {\safeareas} to take advantage of both approaches.
First, the \emph{Proactive Scale Predictor} predicts a coarse-grained estimate of the {\safearea} per each ROI (e.g., 90/70/50/30\% rescaling) to significantly reduce the number of candidate scales to explore.
We identify a set of visual features that closely capture the difficulty of detection and are extractable at minimal overheads (Table~\ref{tb:image-features}) and train a lightweight ML model.
Then, the \emph{Reactive Scale Tuner} further explores a few nearby candidate scales to obtain the optimal {\safearea}.
Due to the processor-independent design, we can easily handle various mobile accelerators.


\vspace{3pt}
{\bf $\bullet$ {\componentThree} (Section~\ref{subsec:scheduler}).} 
Given the rescaled ROIs, our system packs them into high-resolution {\packedframes} to maximize processing throughput.
To this end, we formulate the problem as the combination of \emph{continuous-valued 0-1 knapsack}, \emph{priority-based ordering}, and \emph{2D bin packing} problems to determine (i) the number and sizes of {\packedframes} to maximize utilization of the GPU while satisfying the app latency requirement, (ii) the best processor for each ROI, and (iii) the optimal positioning of ROIs within the {\packedframes} to maximize the fillup ratio.
As the problem is NP-hard, we develop an efficient approximation solution by adopting (i) a branch-and-bound algorithm~\cite{kolesar1967branch} to reduce the search space of the knapsack problem, (ii) latency estimation-based processor selection, and (iii) a Guillotine algorithm~\cite{jylanki2010thousand} to find an effective solution to the 2D bin-packing problem.

We also devise a suite of optimization techniques including packed canvas-aware DNN training (e.g., adding borders around the ROIs to provide hints to the DNN, ground truth filtering), pipelining, consecutive drop minimizing packing policy, and dropped ROI interpolation (Section~\ref{subsec:additional-optimizations}).



Our major contributions are summarized as follows:

\begin{itemize}
    \item We propose {\systemname}, a high-performance edge-enabled object detection system. It adopts a new \textit{\approachname} approach to dynamically scale the ROIs and pack them into large canvases to maximize the parallelism of accelerators. Our approach is applicable to various underlying detection models, accelerators, and inference frameworks.
    \item We systematically explore the effect of ROI scaling on inference accuracy and propose an accurate and efficient ROI scale estimator. This is the first proposal to dynamically rescale ROIs based on the processor and their contents in object detection.
    \item We develop an ROI packing scheduler that intelligently combines the ROIs to large {\packedframes} to maximize the throughput of accelerators while satisfying the latency requirements by considering dynamic capacity of accelerators at runtime.
    \item We perform an extensive evaluation with state-of-the-art accelerators, datasets, and DNNs. Our evaluation shows {\systemname} outperforms the state-of-the-art baselines by 15.0-19.7\% higher mAP, leading to $\times$6.65 higher throughput than frame-wise inference.
\end{itemize}





\section{Motivation}
\label{sec:motiv}

\subsection{Target Scenarios}
\label{subsec:motiv-scenarios}

\begin{figure}[t]
    \centering
    \includegraphics[width=0.9\columnwidth]{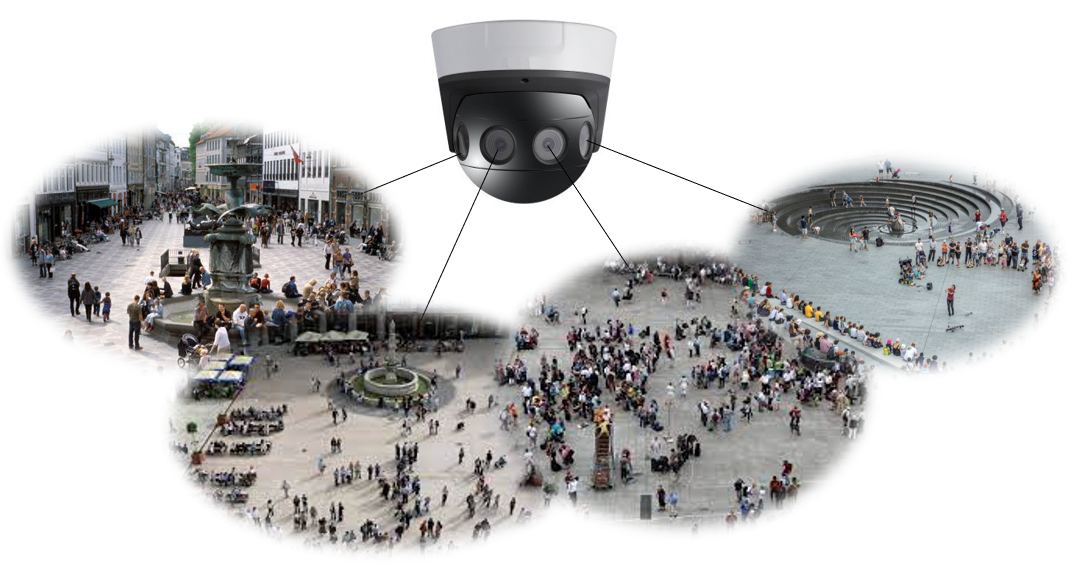}
	\caption{Example scenario of {\systemname}: four-way surveillance camera capturing crowded public square.}
    \label{fig:scenario-figures}
\end{figure}

\noindent \textbf{Political Demonstration.}
The police department monitors various security threats at a political demonstration with large crowds, e.g., counting people in unsafe areas and detecting quickly passing-by criminals or suspicious vehicles. Here, the police would like to (i) install the minimal number of high-resolution camera devices, e.g., one four-way 1080p camera (Figure~\ref{fig:scenario-figures}) installed at the center of a square, (ii) run them at the minimum cost with minimal/zero involvement of clouds, and (iii) process the high resolutions videos at high FPS to identify momentary events from a distance.

\vspace{3pt}
\noindent \textbf{Shoplifting in Supermarket.}
Surveillance cameras monitor aisles in a big supermarket to detect a short moment of shoplifting. Here, a high-resolution camera covers a long aisle from one end. The monitoring should be done at a high frame rate as the shoplifting event may happen instantly (100~ms that spans only three frames in a 30~FPS video). Similarly, high-performance edge-side monitoring can enable this service at a low cost and privacy-preserving manner.

\subsection{Design Goals}

\noindent \textbf{High Throughput.}
Our primary goal is achieving high inference throughput.
This is crucial for processing multiple high-resolution video streams (e.g., 4$\times$1080p@30FPS for the first scenario) and detecting events occurring across only a few frames (e.g., 100~ms for shoplifting).

\vspace{3pt}
\noindent \textbf{On-Device Processing.}
To minimize the deployment and maintenance costs and to deal with privacy concerns, we aim to process the camera streams in an on-device manner without streaming them to the servers.

\vspace{3pt}
\noindent \textbf{Near Real-time Latency.}
Along with high throughput, we also aim to provide soft real-time result delivery (e.g., <5~seconds) to alert the users quickly in case of events.

\subsection{Limitations of Prior Works}

\subsubsection{Lightweight Object Detectors}
\label{subsec:motiv-object-detection-performance}

\begin{table}[t]
\begin{adjustbox}{width=0.99\columnwidth}
\centering
\begin{tabular}{ccccccc}
\toprule
\singlecol{3}{Input\\Resolution} & \multicolumn{3}{c}{\singlecol{2}{Adreno 730\\Galaxy S22 GPU}} & \multicolumn{2}{c}{\singlecol{2}{Hexagon 780\\Galaxy S22 DSP}} & \singlecol{2}{Mali-G78\\Pixel 6 GPU} \\
                                 &         &         &                                             &         &                                                        &                                        \\ \cmidrule{2-7}
                                 & YOLOv5x & YOLOv5m & SSD                                         & YOLOv5x & YOLOv5m                                                & SSD                                    \\ \midrule
320$\times$320                   & 10.77   & 32.36   & 13.74                                       & 108.92  & 378.24                                                 & 12.83                                  \\ \midrule
640$\times$640                   &  4.45   & 13.84   &  5.99                                       &  39.44  & 124.69                                                 & 5.51                                   \\ \midrule
960$\times$960                   &  1.99   &  6.81   &  2.74                                       &  16.51  &  20.36                                                 & 2.64                                   \\ \midrule
1920$\times$1920                 &  0.53   &  1.44   &  0.66                                       &   3.63  &   5.74                                                 & 0.65                                   \\ \bottomrule
\end{tabular}
\end{adjustbox}
\vspace{5pt}
\caption{Input resolution - throughput (FPS) tradeoff of object detectors on various devices.}
\label{tb:lightweight-object-detectors}
\vspace{-5pt}
\end{table}

Lightweight object detectors have been actively proposed, but their throughput is still significantly limited (<2~FPS) for emerging high-resolution cameras.
We investigated the performance of widely-used lightweight object detection models (YOLOv5x, YOLOv5m~\cite{yolov5}, and SSD~\cite{liu2016ssd}) over an 1080p MTA video data~\cite{mta} with three recent accelerators. Table~\ref{tb:lightweight-object-detectors} shows the results. Models with low input resolution achieves high throughput, but lightweight models are incapable of achieving sufficient accuracy with low input resolution (0.238~mAP@0.5 for 320$\times$320 YOLOv5m on Adreno 730 GPU, and 0.280~mAP@0.5 for 640$\times$640 YOLOv5x on Hexagon 780 DSP). When the resolution gets as high as 1920$\times$1920, there is no accuracy degradation, but the throughput drops to <6~FPS. Even with a more powerful edge device, throughput still remains limited (e.g., 3.66~FPS for YOLOv4~\cite{yolov4} model with 800$\times$800 input on Jetson TX2).

\subsubsection {System Optimization Techniques}
\label{subsec:motiv-prior-work-limitations}

A rich body of prior work accelerates object detectors on resource-constrained devices. 
Here, we investigate widely-used approaches: (i) extracting ROIs ~\cite{liu2019edge, zhang2021elf, zeng2020distream}, (ii) tracking \cite{apicharttrisorn2019frugal, flexpatch}, and (iii) utilizing heterogeneous processor~\cite{jeong2022band}.



\vspace{3pt}
\noindent {\bf $\bullet$ ROI-based Approaches.}
The first approach is to process partial regions of the frame since not all contents are of interest.
For example, most of the pixels in a video frame are non-object background (e.g., 80\%~\cite{du2020server}).
While this approach is simple and effective, it has new inefficiencies.

First, ROIs are often larger than the size required by the detector to identify the objects.
For instance, close-by, unoccluded, front-facing, or bright objects are accurately detected with high confidence.
They can be further resized without harming the detection accuracy to improve efficiency.
Recently, Remix utilized a similar opportunity by adapting DNN complexity depending on the detection difficulty of each ROI~\cite{remix}. However, it suffers from low adaptation granularity compared to our ROI rescaling. It requires training multiple backbones (6-7) with various complexity and loading them on memory-constrained edge devices~\cite{lee2020fast}. Also, our system manages ROIs at an object level for fine-granule rescaling while Remix uses course-grained ROIs.

\begin{figure}[t]
    \centering
    \includegraphics[width=0.95\columnwidth]{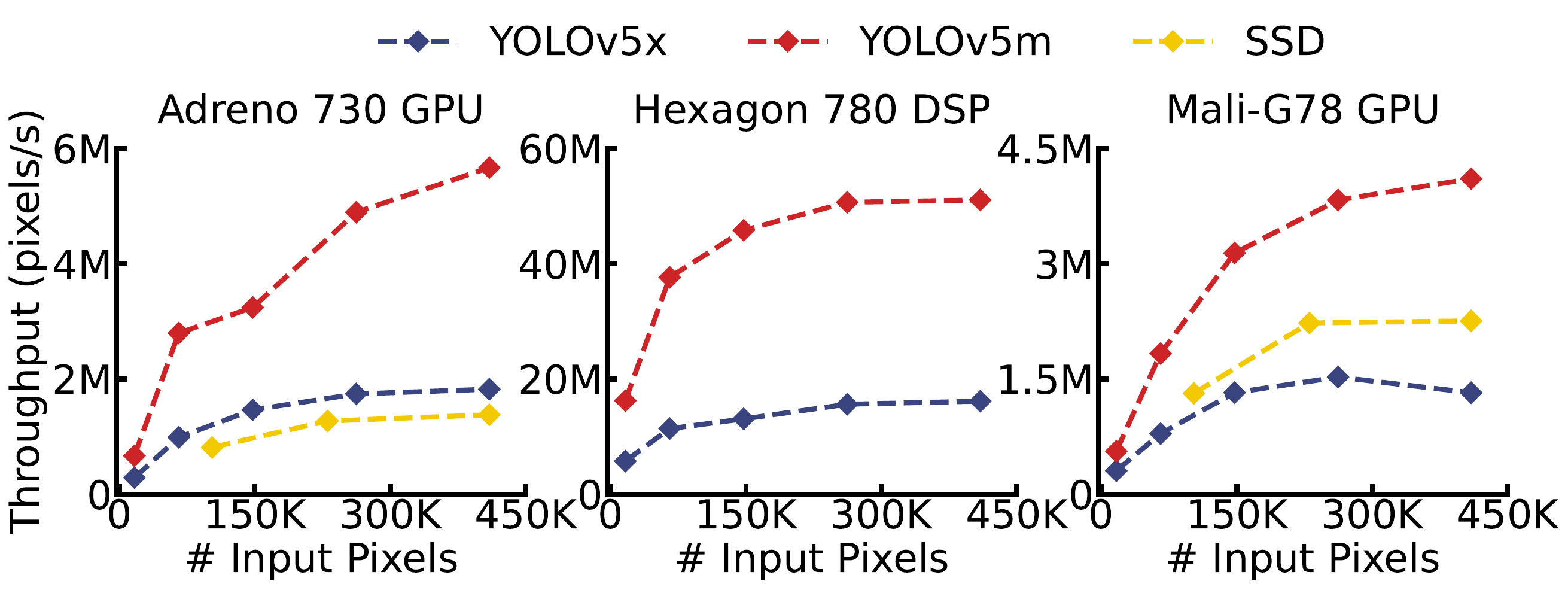}
    \caption{Effect of input size on processing throughput. Pixel throughput means the number of processing pixels in a second.}
    \label{fig:pixel-throughput}
\end{figure}

\begin{figure*}[t]
    \centering
    \begin{subfigure}[t]{0.50\textwidth}
        \centering
        \includegraphics[width=\columnwidth]{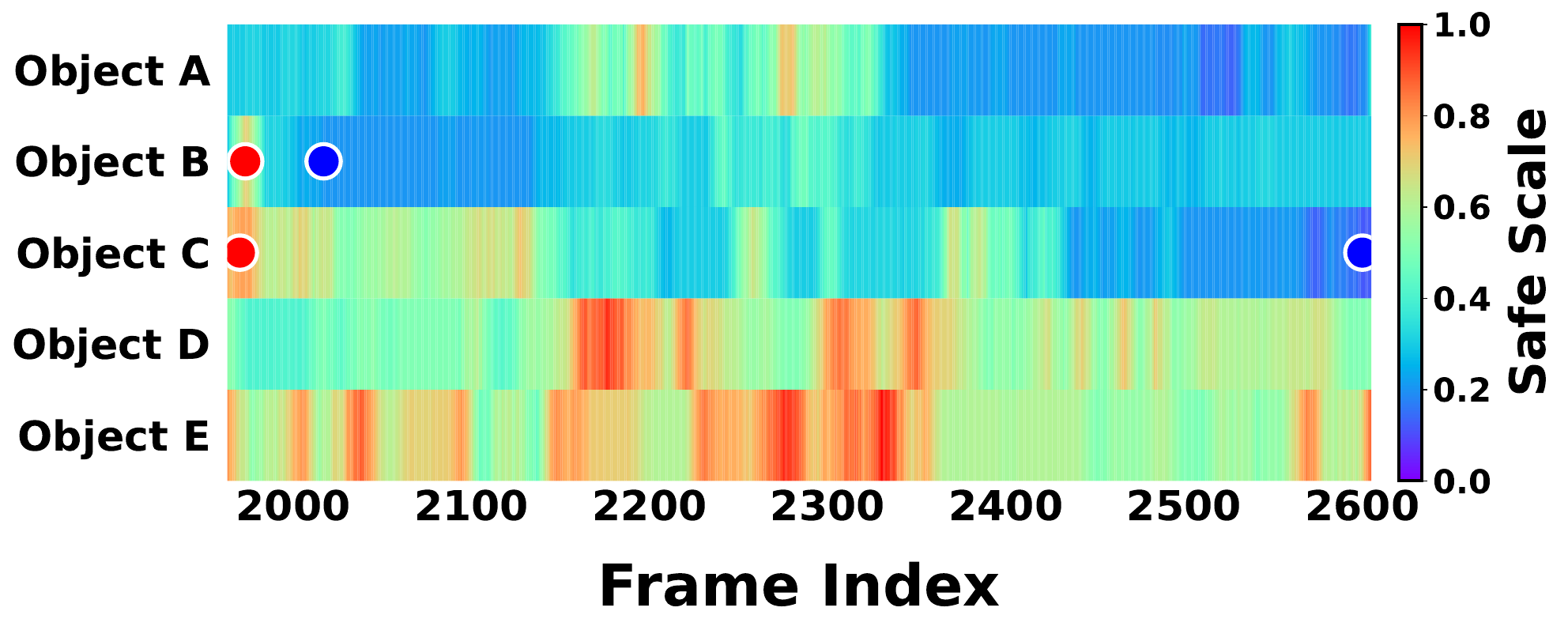}
        \caption{Object-wise {\safeareas} in an MTA video over time.}
        \label{fig:approach-temporal-variation-of-optimal-targets}
    \end{subfigure}
    \begin{subfigure}[t]{0.20\textwidth}
        \centering
        \includegraphics[width=\columnwidth]{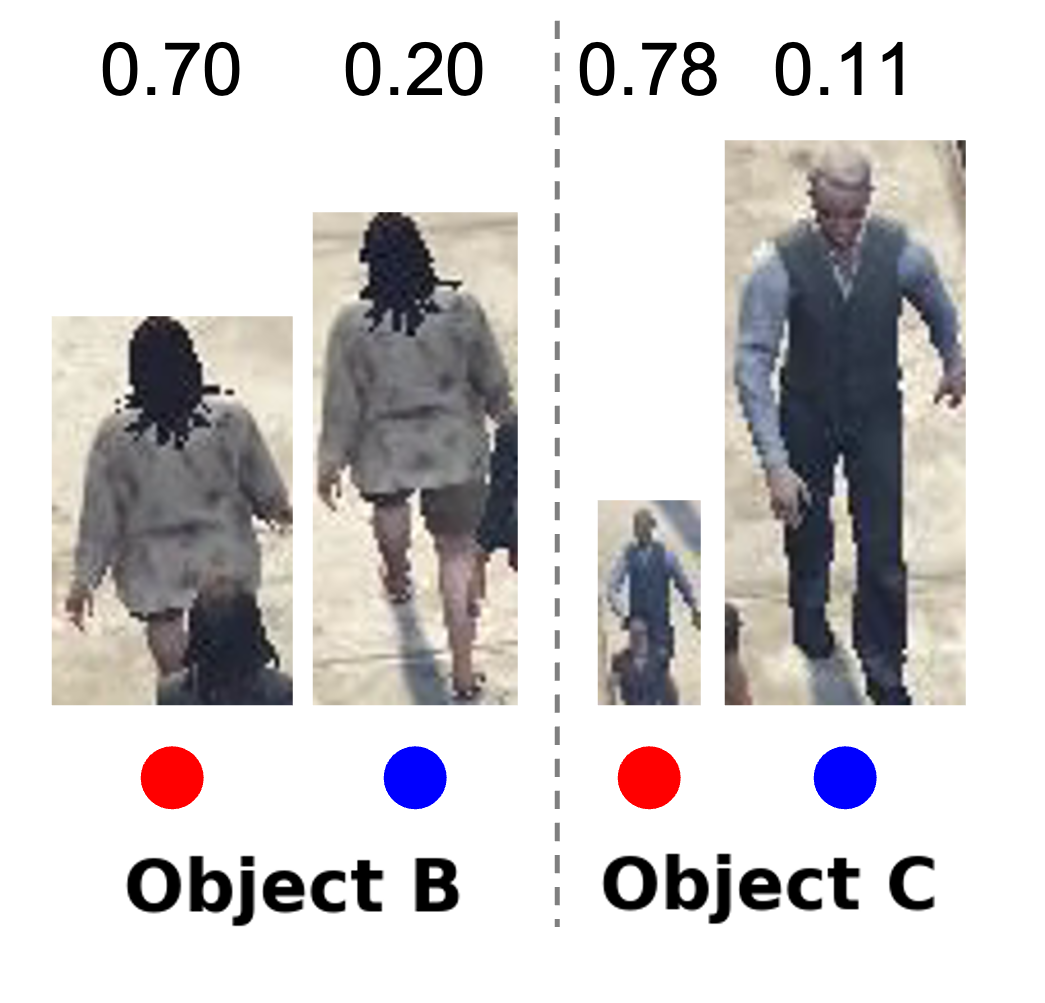}
        \caption{Example ROIs.}
        \label{fig:approach-sample-rois}
    \end{subfigure}
    \begin{subfigure}[t]{0.20\textwidth}
        \centering
        \includegraphics[width=\columnwidth]{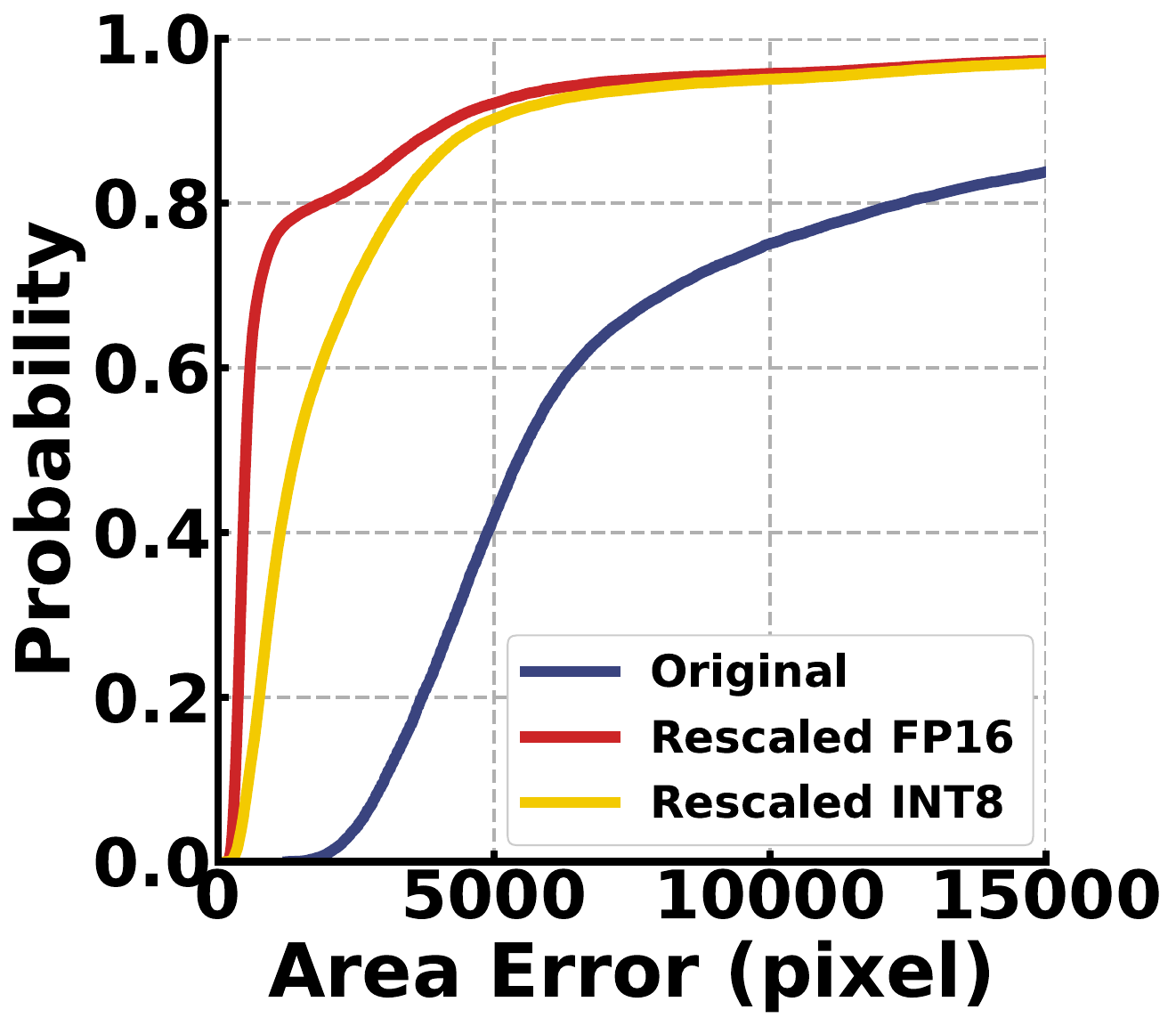}
        \caption{CDF of ROI safe areas. }
        \label{fig:approach-cdf-of-optimal-targets}
    \end{subfigure}
    \caption{Motivational study on the spatio-temporal variation of {\Safearea} (MTA dataset~\cite{mta}).}
    \label{fig:approach-spatio-temporal-safe-area-variation}
\end{figure*}

Second, the input size reduction does not linearly contribute to the throughput gain.
This is mainly because the detectors are processed with accelerators like GPUs and NPUs that provide higher efficiency with larger inputs.
Our evaluation shows that the parallelism of GPU and DSP is severely under-exploited when processing small inputs; Figure~\ref{fig:pixel-throughput} presents the pixel throughput, i.e., the number of pixels processed per unit time increases for the larger inputs.
We may apply batched inference by combining multiple ROIs to improve processing efficiency.
However, many mobile DNN inference frameworks such as TensorFlow-Lite~\cite{tensorflow} do not support batched inference. Even with frameworks that support batched inference (e.g., MNN~\cite{alibabaMNN}), batched inference over variable-size ROIs is not feasible.
We could resize the ROIs to the same pre-defined size to take the benefit of batching, but different ROIs have different optimal target sizes (Figure~\ref{fig:approach-spatio-temporal-safe-area-variation}), making this approach sub-optimal.

\vspace{3pt}
\noindent {\bf $\bullet$ Tracking-based Approaches.}
Another widely-used approach is to track objects across a continuous stream of frames, instead of performing frame-wise detection.
For example, Marlin~\cite{apicharttrisorn2019frugal} efficiently extracts motion vectors of objects and updates the positions of corresponding bounding boxes, improving processing throughput.
FlexPatch~\cite{flexpatch} only runs detection on objects whose tracking accuracy is low.
This approach, however, degrades detection accuracy due to accumulated errors over a series of frames. Such errors quickly propagate when feature points used for tracking are inaccurately extracted due to occlusion or rapid drafts. Also, it fails to detect newly appearing objects, making it hard to apply to dynamic scenes. Recent work aims at overcoming this limitation~\cite{flexpatch}, but it still shows low detection accuracy for dynamic video inputs and needs hand-crafted tuning of various system parameters.

\section{{\rawsystemname} Overview}
\label{sec:appraoch}

\begin{figure}[t]
    \centering
    \includegraphics[width=1\columnwidth]{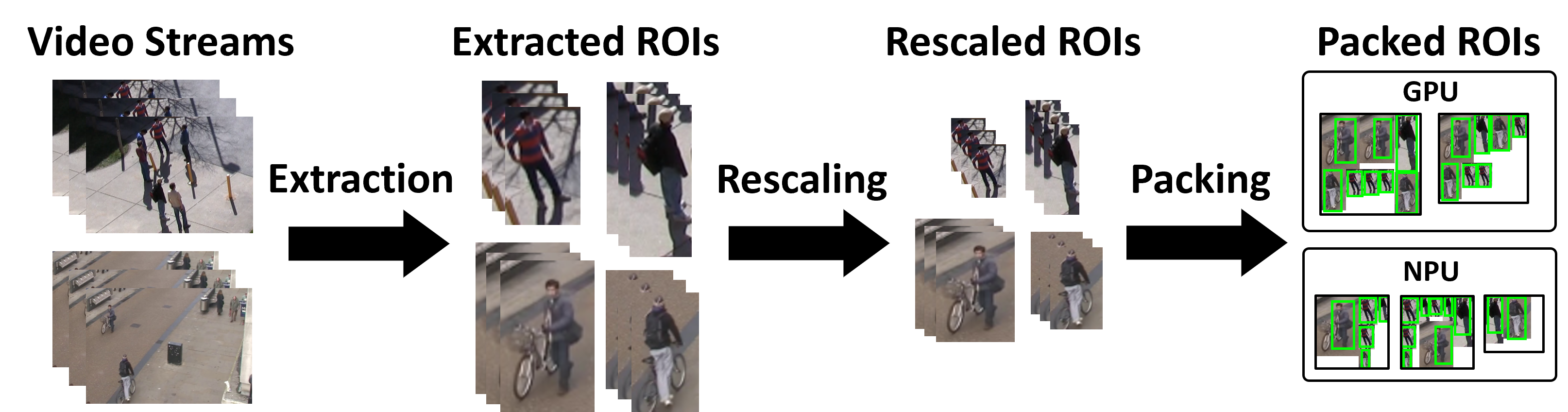}
    \caption{Overview of {\approachname}.}
    \label{fig:approach}
\end{figure}

\subsection{Approach}

We take the \emph{\textit{{\approachname}}} approach to overcome the challenges analyzed in Section~\ref{subsec:motiv-prior-work-limitations}.
Specifically, we achieve both high throughput and accuracy simultaneously through two techniques (Figure~\ref{fig:approach}).

\vspace{3pt}
{\bf $\bullet$ Accuracy-aware ROI Rescaling.} 
We improve the ROI-based techniques to remove unnecessary backgrounds from processing and minimize the number of pixels to process for each ROI.
We achieve this by carefully analyzing the content of each ROI, and aggressively rescaling its size to a minimum without dropping its detection accuracy.

\vspace{3pt}
{\bf $\bullet$ ROI Packing.} 
We maximize the processing parallelism (i.e., number of pixels processed per second) by packing the ROIs into high-resolution images (i.e., canvases) to maximize the pixel throughput (as analyzed in Figure~\ref{fig:pixel-throughput}) with variable-sized scaled ROIs while meeting the latency requirement of the application.
By running the detector on the packed canvases, we fully utilize the parallel computing capabilities of the inference accelerators (e.g., GPU, NPU).

\subsection{Challenges}

\noindent\textbf{High Variation of {\SafeAreas}}.
Determining the {\safearea} is challenging as (i) the {\safearea} varies for different ROIs in the same video frame (depending on its size, whether it is occluded or not, etc.), (ii) the {\safearea} for each ROI changes over time, and (iii) different {\safeareas} of an ROI from heterogeneous accelerators with various precisions.
A single, static scale will either miss the opportunity for throughput improvement (i.e., ROIs remain too large) or incur detection failure (i.e., ROIs are resized too small).
Figure~\ref{fig:approach-temporal-variation-of-optimal-targets} shows how the {\safearea} differs between ROIs and changes over time for ROIs capturing walking people in the MTA dataset~\cite{mta} and the YOLOv5m model. 
It shows that the {\safearea} can either vary within a single second due to a change in visual contents, such as an occlusion in object B (0.70 to 0.20, Figure~\ref{fig:approach-sample-rois} left), or within 10s of seconds due to a change in the size of ROI itself due to navigation as in object C (0.78 to 0.11, Figure~\ref{fig:approach-sample-rois} right).
Figure~\ref{fig:approach-sample-rois} shows an ROI with different {\safeareas} for the quantized YOLOv5m model (for DSP or NPU). The safe areas are larger than those in the original YOLOv5 due to the quantization effect, but still they are much smaller than the original ROI areas.

\vspace{3pt}
\noindent\textbf{{\SafeArea} Profiling Overhead}.
Furthermore, the {\safearea} profiling overhead should be kept at a minimum; otherwise, the overhead will diminish the throughput gain from scaling.
Recent works proposed profiling techniques to determine the optimal video quality (e.g., resolution or quantization parameter) for live video analytics pipeline by running the DNN with different parameters repetitively on the cloud~\cite{jiang2018chameleon, li2020reducto}.
However, this is challenging for {\systemname} in two aspects.
First, we need to run the detector over multiple candidate scales per each object separately (whose total number can be very high in crowded scenes), which incurs significant overhead for resource-constrained edge devices.
Also, the profiled {\safearea} may likely be invalidated quickly for dynamic, high-resolution scenes requiring frequent re-profiling.

\vspace{3pt}
\noindent\textbf{Efficient ROI Packing}.
The shapes and sizes of the ROIs are highly irregular 
(e.g., the average size and standard deviation of the extracted ROIs in MTA dataset~\cite{mta} are  9,355 and 9,390, and those of rescaled ROIs are 1,128 and 1,877 pixels (Figure~\ref{fig:approach-cdf-of-optimal-targets}).
Na{\"i}vely planning the high-resolution frames and packing ROIs into them incurs significant inefficiency.
Finding an optimal grouping solution is challenging, as the search space is O(N!) for N ROI sequences.

\begin{figure}[t]
    \centering
    \includegraphics[width=1\columnwidth]{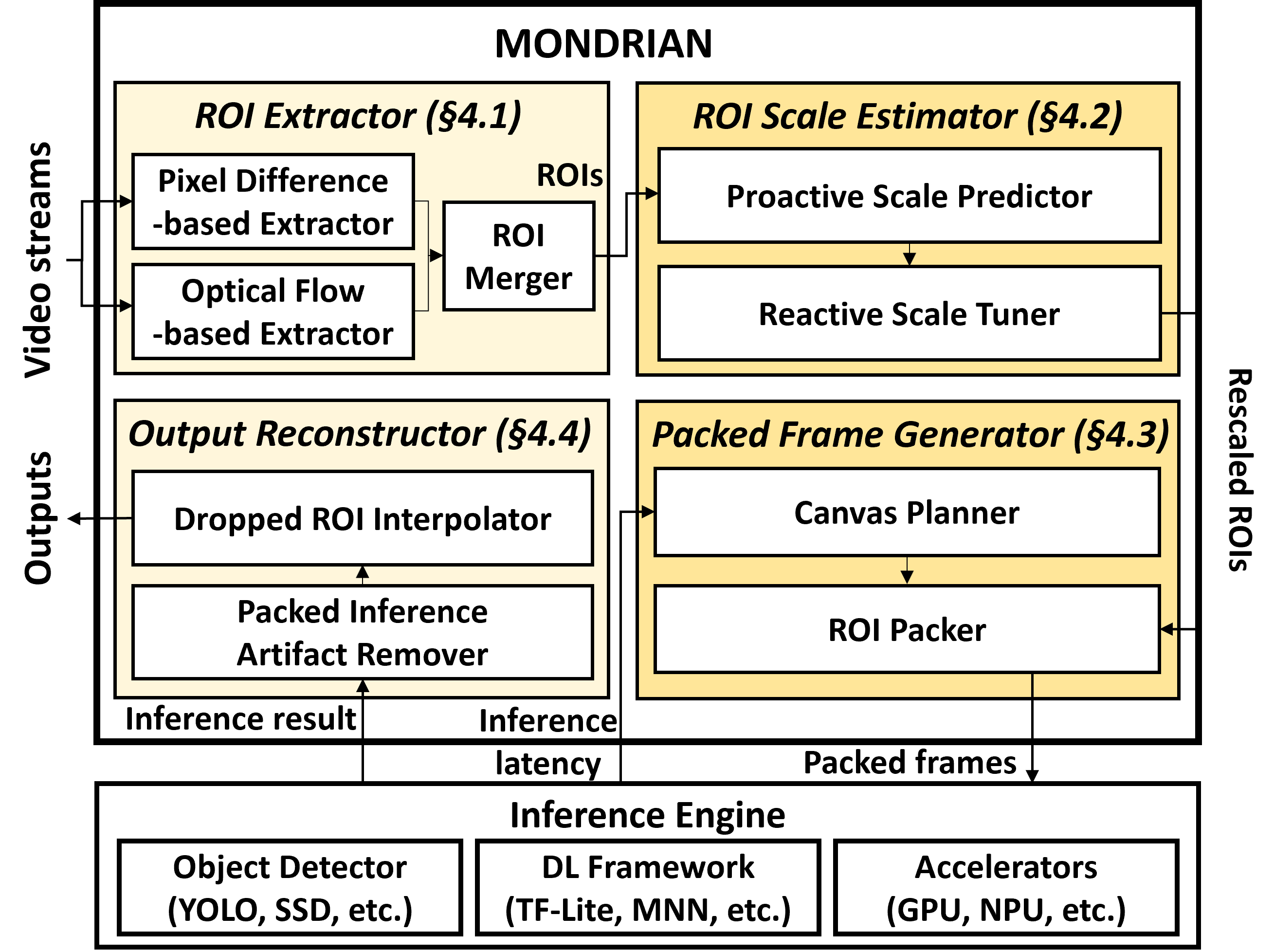}
    \caption{System architecture of {\systemname}.}
    \label{fig:system-architecture}
\end{figure}

\subsection{System Architecture}

Figure~\ref{fig:system-architecture} shows the overall architecture of {\systemname}.
 \textit{Lightweight ROI Extractor} first extracts the ROIs from video streams efficiently identifying newly appearing objects as well as tracking already detected objects (Section~\ref{subsec:roi-extractor}).
Then, \textit{\componentTwo} rescales the extracted ROIs to the {\safearea}; the {\safearea} is tracked in a per-ROI manner through a hybrid approach (lightweight ML-based prediction and reactive probing-based calibration) (Section~\ref{subsec:scale-estimator}).
Third, \textit{\componentThree} groups and packs the scaled ROIs into {\packedframes} to maximize the processing parallelism (Section~\ref{subsec:scheduler}).

After running the detector inference on the generated {\packedframes}, the results are reconstructed back to the original frame in the {\packedinference} process~(Section~\ref{subsec:scheduler-inference}).
Finally, {\systemname} is composed of multiple optimization techniques including model fine-tuning on packed frames, dropped ROI interpolation, and pipelining (Section~\ref{subsec:additional-optimizations}).

\section{{\rawsystemname} Inference Pipeline}

\subsection{{\componentOne}}
\label{subsec:roi-extractor}
The first stage of the {\systemname} inference pipeline is ROI extraction.
ROI extraction has been widely studied in the context of computation offloading to reduce the data transfer \cite{liu2019edge, du2020server}.
However, few works studied an ROI extraction method in the context of on-device processing, mainly because the insufficient area of input degrades pixel efficiency (i.e., the number of pixels processed per second) in accelerators, failing to enhance the performance effectively. While we acknowledge that ROI extraction is not conceptually new, we carefully put together and specialize existing algorithms for stable and lightweight ROI extraction.

\vspace{3pt}
\noindent {\bf ROI Extractors.}
We employ two lightweight ROI extractors: (i) Optical Flow (OF)-based ROI extractor to track existing objects, and (ii) Pixel Difference (PD)-based ROI extractor to identify newly appearing objects, and merges their results for subsequent ROI rescaling and packing.

{\bf $\bullet$ OF-based ROI Extractor} propagates the bounding boxes detected by the object detector over subsequent frames by an object tracking algorithm \cite{lucas1981iterative}.
Object tracking comprises (i) feature point selection and (ii) the optical flow estimation stages.
We use Shi-Tomasi~\cite{shi1994good} for feature point extraction and Lucas-Kanade~\cite{lucas1981iterative} for optical flow estimation.
The two stages take a total of 13~ms on the Galaxy S22 smartphone for a single FHD frame. 
Note that we use tracking only for ROI proposals (not for skipping detector inferences), as tracking fails in dynamic scenes (Section~\ref{subsec:motiv-prior-work-limitations}).

{\bf $\bullet$ PD-based ROI Extractor} handles object that newly appears or is unstably tracked by the OF-based extractor due to abrupt changes in object poses, occlusion, etc.
It operates in four stages: (i) take the pixel-wise difference between two frames of small gap (e.g., 5 frames), (ii) remove noises by ignoring small differences, (iii) detect edge and contour using lightweight Canny edge detector~\cite{canny1986computational} and Border Following algorithm~\cite{suzuki1985topological}, and (iv) extract minimum bounding box surrounding the contours as ROIs. 
The PD-based ROI extractor runs at 5.23~ms on Galaxy S22 for an FHD frame.

\vspace{3pt}
\noindent {\bf Post Processing.}
We merge overlapping ROIs suggested by PD-based and OF-based ROI extractors to remove redundancy.
Specifically, we consider two ROIs with Intersections over union (IoU) > 0.5 to be overlapped.
Also, we pass the various visual features of ROIs (e.g., motion vectors and tracking error for OF-tracked ROIs) to the ROI scale estimator.

\vspace{3pt}
\noindent {\bf ROI Merging.}
We merge overlapping ROIs from the PD- and OF-based extractors to reduce compute redundancy.
Specifically, we merge two ROIs if two conditions are satisfied.
(i) IoU is larger than a threshold (e.g., >0.8).
(ii) The rescaled area of the merged ROI is smaller than the sum of rescaled areas of individual ROIs.

\begin{figure}[t]
    \begin{center}
        \includegraphics[width=1\columnwidth]{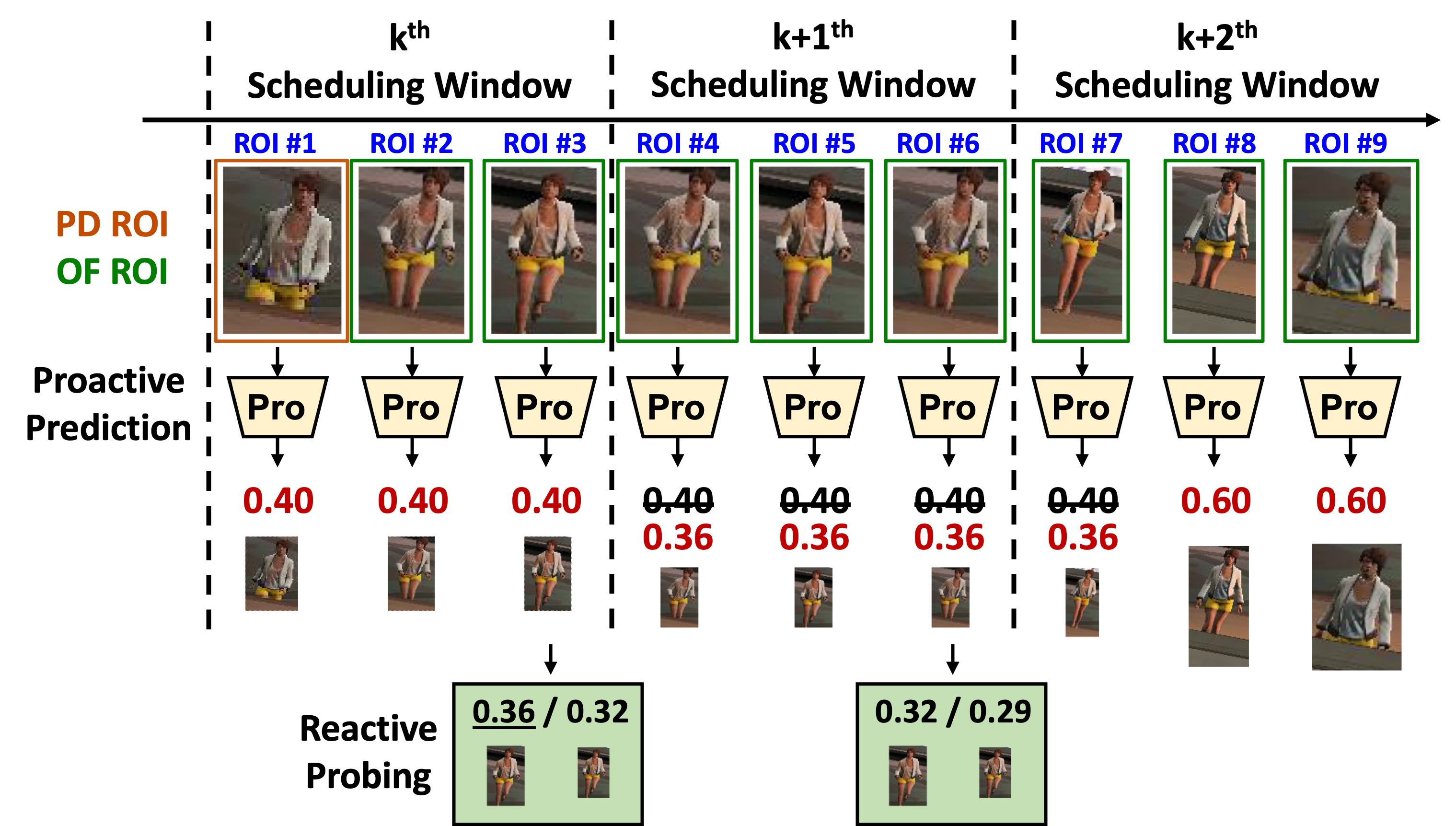}
        \caption{Operational flow of ROI Scale Estimator (red values indicate the estimated {\safeareas}). }
        \label{fig:system-reactive-probing}
    \end{center}
\end{figure}

\subsection{{\componentTwo}}
\label{subsec:scale-estimator}
 
The goal of \textit{\componentTwo} is to accurately estimate the time-varying {\safeareas} for each ROI at a low overhead. 
There are two possible approaches: (i) \emph{proactive prediction} (i.e., estimating the {\safearea} from lightweight input features)~\cite{chin2019adascale, xu2020approxdet, xu2022litereconfig} and (ii) \emph{reactive probing} (i.e., running the DNN inference over multiple rescaled ROIs)~\cite{jiang2018chameleon, zhang2018awstream}. 
The two approaches have pros and cons.
It is hard to achieve high accuracy with prediction, mainly due to difficulty of DNN accuracy estimation (works only for simple content at a coarse-grained scale~\cite{chin2019adascale}).
Probing is accurate but incurs high overhead (e.g., rescale ratio ranges from 1$\times$ to 0.1$\times$ in MTA dataset~\cite{mta}).

We take a hybrid approach to address the challenges.
First, \textit{\estimatorOne} (Section~\ref{subsubsec:scale-predictor}) estimates the resize target at a coarse-grained level with minimal overhead.
In particular, it leverages zero-cost features that can be obtained from the ROI extractor and the detector inference results (e.g., optical flow magnitude, aspect ratio, optical flow tracking error, etc.) and applies a lightweight machine learning model for estimation. 
Second, \textit{\estimatorTwo} (Section~\ref{subsubsec:scale-tuner}) incrementally performs finer adjustments towards optimal resize scale by selectively runs the inference on a couple of {\safearea} candidates near the coarse-grained prediction.

\subsubsection{Operational Flow}
\label{subsubsec:scale-estimator-overview}
Figure~\ref{fig:system-reactive-probing} depicts the operational flow of {\componentTwo} for a single ROI.
For the PD-based ROI (ROI \#1 indicating a newly appearing object), we first run the {\estimatorOne} to estimate its coarse target scale in 3 levels: 0.4, 0.6, and 0.8. In the example, the target scale is estimated at 0.4. For the subsequent OF-based ROIs (ROI \#2 onwards indicating tracked objects), the system continues to estimate the target scales with both proactive estimation and reactive tuning. In particular, it performs proactive estimation per every ROI and directly uses the results if the finer adjustment is not available (ROI \#1-\#3 in the figure) or invalidated with substantial changes in proactive estimation (ROI \#8), indicating considerable changes of {\safeareas}. Otherwise, it applies the finer adjustments suggested by the reactive scale tuner (ROI \#4--7). Specifically, the tuner searches for a few nearby smaller scales (e.g., 0.36 and 0.32 if the prediction is 0.4) to check whether more aggressive rescaling is possible. 
Note that reactive tuning is performed once per each scheduling window (not per each frame) to reduce overhead.

\subsubsection{\estimatorOne}
\label{subsubsec:scale-predictor}
We take a data-driven approach to design the lightweight proactive scale estimator. We noticed that {\safearea} prediction is a simpler task compared to the detection itself (that the ML approach works well), even more simpler when the granularity of prediction is coarse. Here, we focus on making it as light as possible for two reasons: i) we need to apply it on a large number of ROIs (e.g., 10-100), and ii) further tuning is done to correct potential errors.
Specifically, we identify a suite of zero-cost features that characterize the detection difficulty of an ROI. Then, we train a lightweight Decision Tree that predicts the target scales using the features; target scale estimation granularity and values are set as quantiles from the {\safearea} distribution in the dataset (e.g., in 5 levels - $\leq$369, $\leq$459, $\leq$559, $\leq$848, $>$848 (pixels) for MTA dataset~\cite{mta}). The predictor takes less than 1~ms to process hundreds of ROIs, effectively reducing the search space for {\safearea} estimation~(Section~\ref{subsubsec:eval-scale-estimator}).

\vspace{3pt}
\noindent{\bf List of Features.}
We carefully categorize the factors that affect the detection difficulty, and identify the features extractable in the ROI extractor at minimal overheads (note that prior works have only utilized simple features such as object size~\cite{remix, chin2019adascale}):

\begin{table}[t]
\renewcommand{\arraystretch}{1.5} 
\large
\centering
\caption{Feature list and trained importances (MTA). }
\begin{adjustbox}{width=0.99\columnwidth}
\begin{tabular}{clc}
\toprule
\singlecol{2}{Feature\\Category} & \singlecol{2}{Feature}                               & \singlecol{2}{Trained Feature\\Importance (\%)} \\
                                 &                                                      &                                                 \\ \midrule
\singlecol{2}{Object\\Size}      & ROI Area                                             & 32.8                                            \\ \cmidrule{2-3}
                                 & ROI Max edge length                                  & 17.5                                            \\ \midrule
\singlecol{2}{Pose \&\\Movement} & ROI Width / height ratio                             & 15.3                                            \\ \cmidrule{2-3}
                                 & Average feature point shifts                         & 9.3                                             \\ \midrule
\singlecol{3}{ROI\\Reliability}  & Average optical flow tracking error                  & 8.9                                             \\ \cmidrule{2-3}
                                 & Standard deviation of feature point shifts           & 8.8                                             \\ \cmidrule{2-3}
                                 & Normalized cross correlation of feature point shifts & 7.5                                             \\ \bottomrule
\end{tabular}
\end{adjustbox}
\vspace{3pt}
\vspace{-10pt}
\label{tb:image-features}
\end{table}

\vspace{3pt}
{\bf $\bullet$ Object Appearance-related Features.}
An object's detection difficulty varies depending on type, size, pose, and movement.
We utilize the ROI size (area and maximum edge length) and aspect ratio as the proxies for size and pose.
Also, we adopt the average magnitude of optical flow shifts as the proxy for movement.
Note that all features are quickly generated in the ROI extraction process.

\vspace{3pt}
{\bf $\bullet$ ROI Reliability-related Features.} 
For OF-based ROIs generated for tracked objects, we leverage additional features that indicates the reliability of ROIs.
Our hypothesis is that ROI reliability indicates the difficulty of detection, which helps to determine the {\safearea}.
For this purpose, we use optical flow error and optical flow shift statistics (standard deviation of shifts, normalized cross correlation of shifts) as the proxies to ROI reliability~\cite{apicharttrisorn2019frugal}. 
Note that these features are available at free of costs from the ROI extraction process.

\vspace{3pt}
\noindent {\bf Predictor Training.}
The training is done offline over the target video dataset and the model; note that the predictor needs to be trained with the INT8-quantized model for DSP and NPU execution while the original FP32 model can be used for GPU. In particular, we generate the above features by performing the ROI extraction process for the training video. 
Also, we label the ground truth of the {\safearea} by repeatedly performing the detection over re-scaled input images with various scale factors and identifying the smallest scale where the detection result is intact. 
We added a small margin (e.g., 0.1) to the ground truth {\safearea} labels, so that the predictor is trained to output conservative predictions (as overly scaling the ROI will result in detection failure and degrades accuracy).
Note that this can be further tuned by the reactive tuner (Section~\ref{subsubsec:scale-tuner}). 

Table~\ref{tb:image-features} lists the features along with their importance scores; the scores are calculated based on how much a feature contributes to reducing the training error in training the Decision Tree classifier.

\subsubsection{\estimatorTwo}
\label{subsubsec:scale-tuner}

Given the coarse estimate of the target scale from the Proactive Scale Predictor, the Reactive Scale Tuner probes the nearby scales to fine-tune the {\safearea} estimation. 
To minimize probing overhead, we design an incremental probing method, which probes only a few nearby options at a step.

\vspace{3pt}
\noindent {\bf Probing Interval.}
For each ROI the probing is periodically triggered once per each scheduling event  (e.g., 1-5 seconds based on the app latency requirement, instead of per every frame) to reduce the overheads. 
Such duty cycling is possible for two reasons: (i) the fine-tuned results remain constant until there are considerable scene content changes, 
and (ii) we already have coarse {\safearea} estimation, preventing the processing of unnecessarily large ROIs. 
Among the frames in a scheduling window, we select the last frame for probing, as the profiled value will be most likely remain consistent for the longest duration in the subsequent window.

\vspace{3pt}
\noindent {\bf Probe Target Scales.}
We probe {\safeareas} smaller than that output by the proactive predictor, as it is trained to output conservative estimates. 
Per each scheduling interval, we empirically probe 5 candidate scales (100\%, 90\%, 80\%, 70\%, 60\% of the {\safearea} suggested by the proactive predictor); see Section~\ref{subsubsec:eval-scale-estimator} for detailed evaluation on the number of probe scales.
Note that the probing scales are set relative to the current scale, with the intuition that the scale must be tuned more delicately at a smaller range.

\vspace{3pt}
\noindent {\bf Probe Output Interpretation.} 
We choose the smallest {\safearea} with successful detection.
We determine that a probe detection result of a scaled ROI is successful if its confidence is higher than 0.3 and IoU with the reference bounding box (the ROI without further re-scaling) is higher than 0.75. 
If all the candidates fail to detect, we calibrate the target scale one step larger from the start scale.

\subsection{\componentThree}
\label{subsec:scheduler}

{\componentThree} maximizes the processing parallelism while maintaining accuracy by packing the set of scaled ROIs into large, high-resolution images (i.e., {\packedframes}).
Achieving the goal incurs two challenges.
(i) \emph{How to determine the number and sizes of the canvases to meet the app latency requirement as well as maximally utilize the parallel processing capability of target processors?}
(ii) \emph{How to efficiently pack the ROIs into the planned canvases to minimize empty spaces?} ROI packing is an NP-hard problem; naively packing the ROIs in their extraction order results in significant inefficiency since the re-scaled ROI sizes are highly variable.
Simultaneously satisfying the two conditions is non-trivial: larger canvases are beneficial for compute efficiency, but have higher chance of latency violation.
While some works have explored similar idea~\cite{flexpatch, kumar2019pack}, they use a single, fixed canvas size; they cannot adapt to runtime inference latency fluctuation due to resource dynamics.

\subsubsection{{\Packedframe} Planning}
\label{subsubsec:scheduler-planning}

Given $k$ square canvas templates with different sizes (e.g., 320$\times$320,  640$\times$640, ..., 1280$\times$1280), our goal is to find how many of each template to use (i.e., $\{x_1, x_2, ..., x_k\}$, where $x_i$ is 0 or a positive integer) in a given scheduling interval (i.e., the app latency deadline $T$). The maximum template size is chosen as the smallest size that fully utilizes processor parallelism, which is profiled offline (examples of three processors are in Figure~\ref{fig:pixel-throughput}); for example, it is chosen as 1280$\times$1280 for the Adreno 730 GPU and 768$\times$768 for the Hexagon 780 DSP). Smaller size canvases are also used to handle cases where the number of ROIs is smaller or the app latency requirement is tight.

We use the following optimization policy to determine $\{x_1, x_2, ..., x_k\}$, which maximizes the total area of the canvases (and throughput).

\begin{equation}
\centering
\small
\begin{aligned}
\text{maximize} \mspace{5mu} \sum_{i=1}^{k} a_i x_i \mspace{10mu} \text{subject to} \mspace{5mu} \sum_{i=1}^{k} l_i x_i \leq T, \mspace{10mu} 0 \leq x_i \in \mathbb{Z},
\end{aligned}
\label{eq:group}
\end{equation}

\noindent where $A = \{a_1, a_2, ..., a_k\}$ are the areas of the candidate canvases and $\{l_1, l_2, ..., l_k\}$ are the inference latencies. 
This problem is equivalent to the \textit{unbounded 0-1 knapsack problem} with continuous constraint.
We adopt a lightweight \textit{branch and bound} algorithm~\cite{kolesar1967branch} to efficiently explore the candidate canvas combinations, whose latency overhead is marginal (around 0.06 ms).

The inference latencies $\{l_1, l_2, ..., l_k\}$ are profiled offline and updated at runtime upon inference finish, to track dynamic resource availability (e.g., thermal throttling).
For runtime latency update, we apply exponential smoothing for the recently used canvas templates and interpolate the latency for the recently unused templates.

\subsubsection{ROI Packing}
\label{subsubsec:scheduler-packing}

The goal of this step is to decide which canvas and where in the canvas to place each ROI to maximize the packing efficiency. This can be formulated as the \emph{N-way 2D bin packing} (i.e., packing a set of 2D re-scaled ROIs into $N$ large canvases (bins), with minimal wasted areas).
As this is a well-known NP-Hard problem~\cite{hartmanis1982computers}, we devise an approximate algorithm based on the Guillotine algorithm~\cite{jylanki2010thousand}. 
In particular, we first sort the extracted ROIs by their priority (will be introduced shortly) and feed them into our algorithm. The algorithm determines (i) the canvas whose remaining area after packing is minimized and (ii) the position of the ROI in the chosen canvas via the Guillotine algorithm. 


\vspace{3pt}
\noindent {\bf ROI Priority Assignment.}
We determine the ROI packing order as follows.
First, we prioritize two types of ROIs and place them in the first canvas: (i) the probe ROIs (to apply the adjustment as fast as possible) and (ii) the ROIs in the last frame in a scheduling interval (to reset the reference positions of the OF-based ROI extractor with the up-to-date detection results).
Next, we pack PD-based ROIs (for new objects) and OF-based
ROIs with low confidence.
In case the remaining ROIs cannot be packed in the planned canvas, we drop them and interpolate the inference results (Section~\ref{subsec:additional-optimizations}).
Within the same type, the ROIs are packed in the descending order of their sizes.

\begin{figure}[t]
    \centering
    \begin{subfigure}[t]{0.28\columnwidth}
        \centering
        \includegraphics[width=\textwidth]{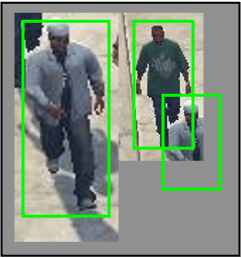}
        \caption{Packed canvas and box labels.}
        \label{fig:packing-preprocess-simple}
    \end{subfigure}
    \begin{subfigure}[t]{0.28\columnwidth}
        \centering
        \includegraphics[width=\textwidth]{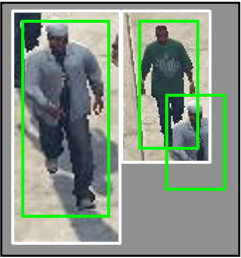}
        \caption{Add border around the ROIs.}
        \label{fig:packing-preprocess-border}
    \end{subfigure}
    \begin{subfigure}[t]{0.28\columnwidth}
        \centering
        \includegraphics[width=\textwidth]{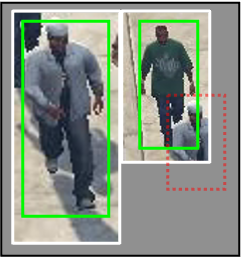}
        \caption{Remove non-target labels (red).}
        \label{fig:packing-preprocess-filter}
    \end{subfigure}
    \caption{Preprocessing for packed canvas fine-tuning.}
    \label{fig:preprocess-fine-tuning}
\end{figure}

\begin{figure}[t]
    \centering
    \includegraphics[width=0.9\columnwidth]{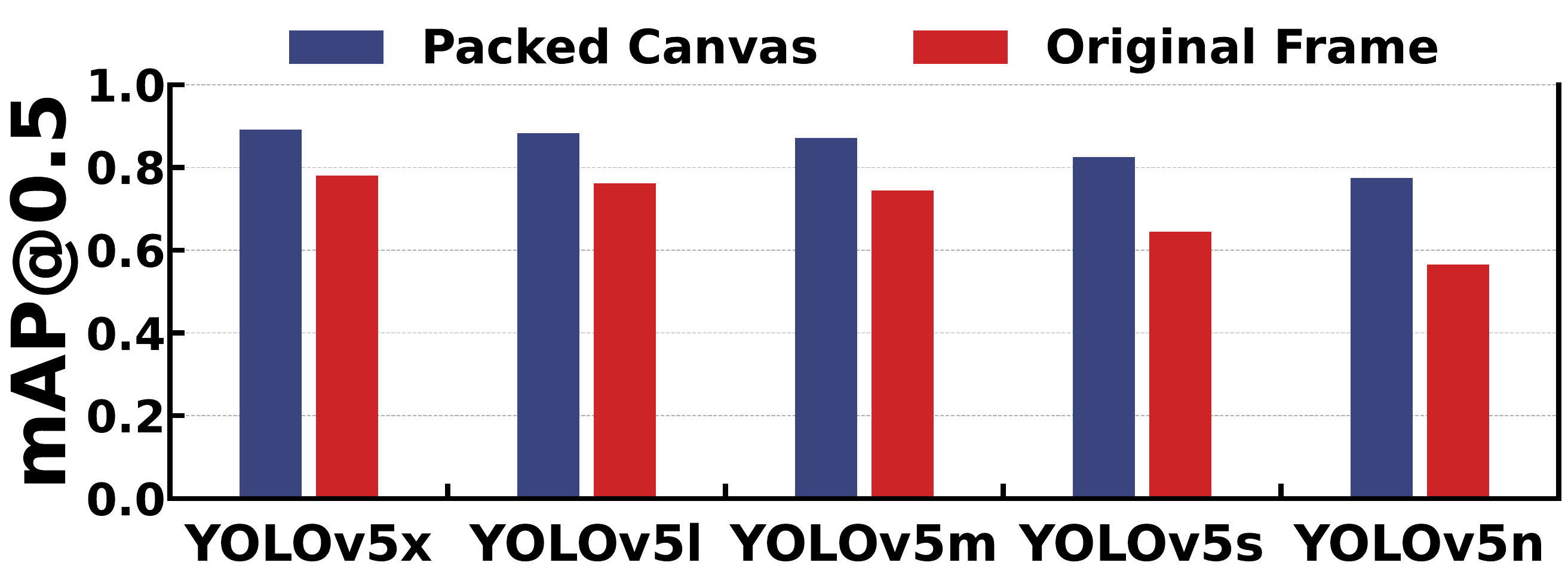}
    \caption{Detection accuracy comparison on packed {\packedframes} vs. original frames (MTA). }
    \label{fig:robust-fine-tuning}
\end{figure}

\subsection{Inference and Output Reconstruction}
\label{subsec:scheduler-inference}

Lastly, we run the detector inference over the generated canvases and convert the detected boxes to the original ROI coordinates. Note that {\systemname} can be integrated with any DNN inference engine (e.g., TensorFlow Lite, MNN).


A single object may be duplicately included in multiple ROIs when two objects are overlapping with each other.
In such cases, the final output would yield duplicate bounding boxes of the same object.
We filter out such artifacts by applying additional non-maximum suppression (NMS) on the reconstructed boxes.

\subsection{Additional Optimizations}
\label{subsec:additional-optimizations}

\noindent {\bf Packed Canvas-aware DNN Fine-Tuning.}
Packed canvases appear different compared to normal images used for object detector training. To improve accuracy, we fine-tune the model on packed frames.
We add borders around the ROIs, which indirectly limits the receptive field (i.e., how large an area the model analyzes to detect objects) within each ROI (Figure~\ref{fig:packing-preprocess-border}).
For ground truth, we only label boxes in the centers of the ROIs, so as to reduce cases where the model detects non-target objects that may be partially included in the ROI extraction process (Figure~\ref{fig:packing-preprocess-filter}).
YOLOv5 n/s/m/l/x fine-tuned on {\packedframes} achieves on average 0.106 mAP gain on packed frames generated from the MTA dataset.
We also observe that the accuracy gain from {\packedframes} fine-tuning compared to full-frame inference increases for lightweight DNNs as shown in Figure~\ref{fig:robust-fine-tuning}.
We conjecture that this is because object detection on {\packedframes} is an easier task compared to that on the full frames (ROIs are clearly separated and the object is mostly placed in the center of each ROI), and even lightweight DNNs can be trained to achieve high accuracy.





\vspace{3pt}
\noindent {\bf Pipelining}
We optimize the inference pipeline by running parallelizable components across multi-core CPU and GPU to hide their overheads. We set the pipeline unit as a frame. Once the ROIs are rescaled, packed, and enqueued for inference on the GPU, subsequent frame processing is done on the CPU in parallel so that we can maintain the GPU utilization high (e.g., 92.5\% as shown in Table~\ref{tb:eval-latency-breakdown}).

\vspace{3pt}
\noindent {\bf Interpolation for Dropped ROIs.}
In case an ROI is dropped from processing in the ROI packing process (Section~\ref{subsubsec:scheduler-packing}), we interpolate its detection result from the detected bounding boxes of the neighboring frames.
Since interpolation does not work well with long consecutive drops, we minimized the length of consecutive drops by packing ROIs with the longest consecutive drops first.




\section{Evaluation}

\begin{figure}[t]
    \centering
    \begin{subfigure}[t]{0.55\columnwidth}
        \centering
        \includegraphics[width=\textwidth]{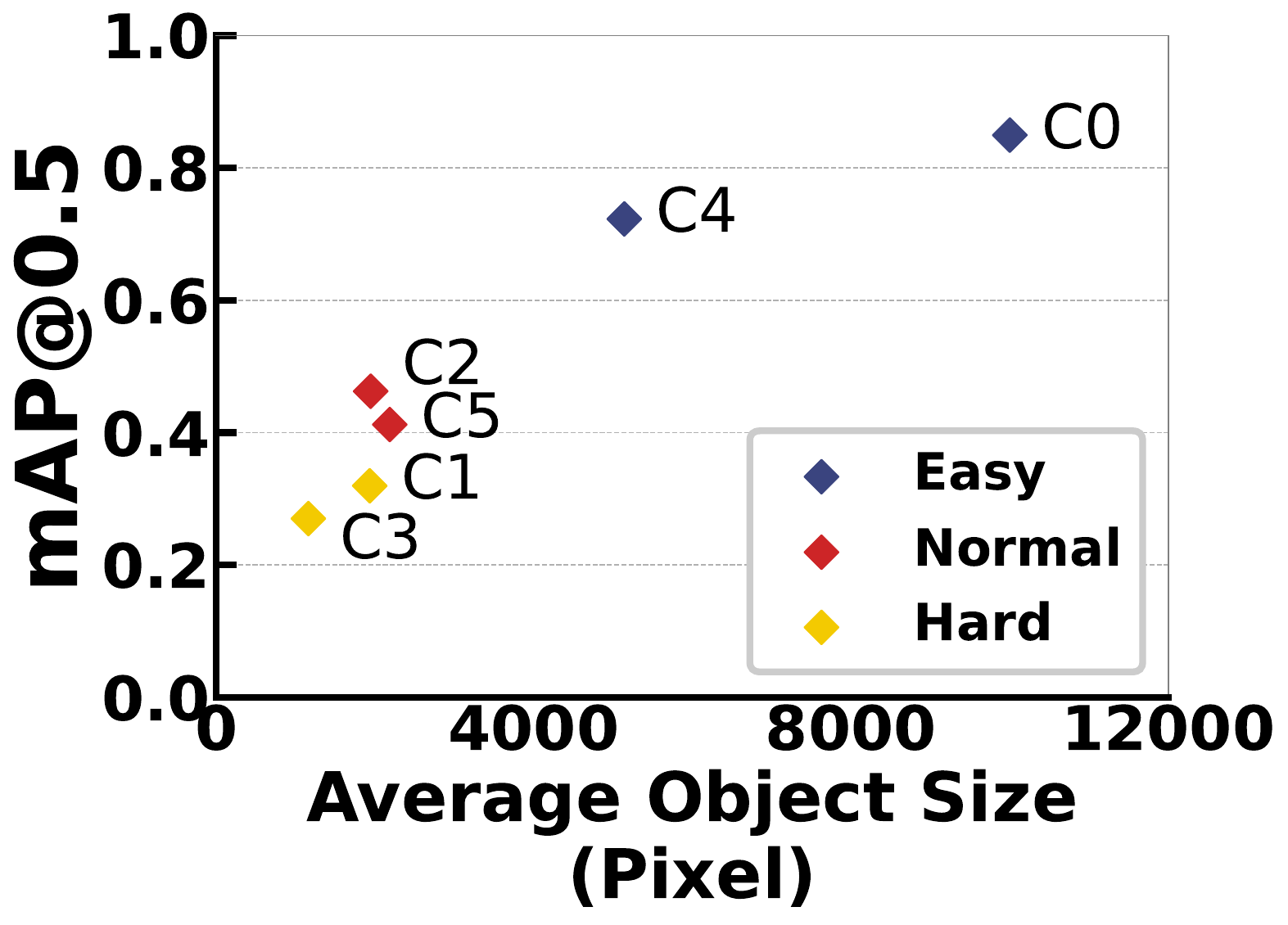}
        \caption{Difficulty of each video.}
    \end{subfigure}
    \begin{subfigure}[t]{0.43\columnwidth}
        \centering
        \includegraphics[width=\textwidth]{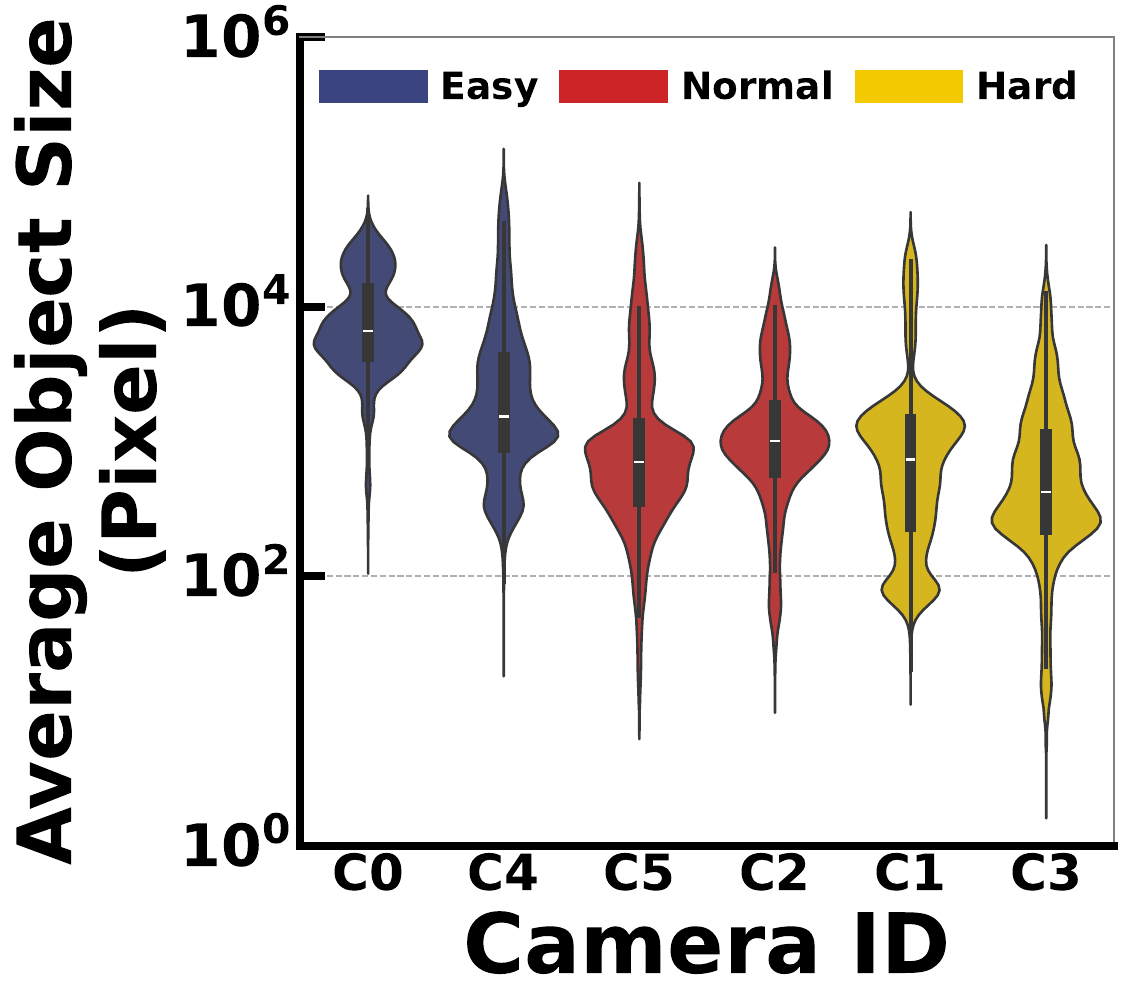}
        \caption{Size distribution.}
    \end{subfigure}
    \caption{Video grouping by difficulty.}
    \label{fig:eval-video-info-stats}
\end{figure}

\begin{figure}[t]
    \centering
    \begin{subfigure}[t]{0.32\columnwidth}
        \centering
        \includegraphics[width=\textwidth]{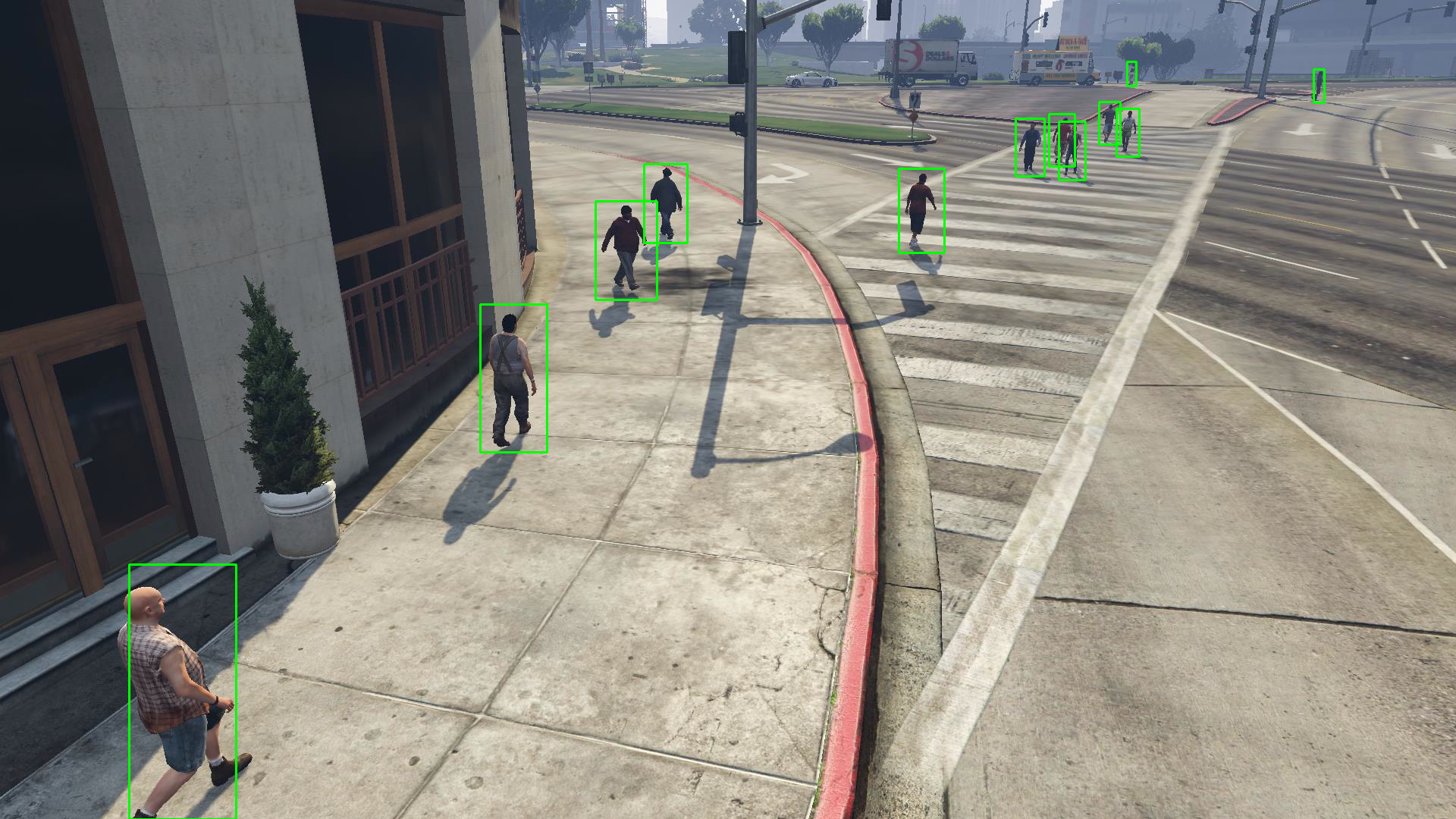}
        \caption{Easy: CAM 4}
    \end{subfigure}
    \begin{subfigure}[t]{0.32\columnwidth}
        \centering
        \includegraphics[width=\textwidth]{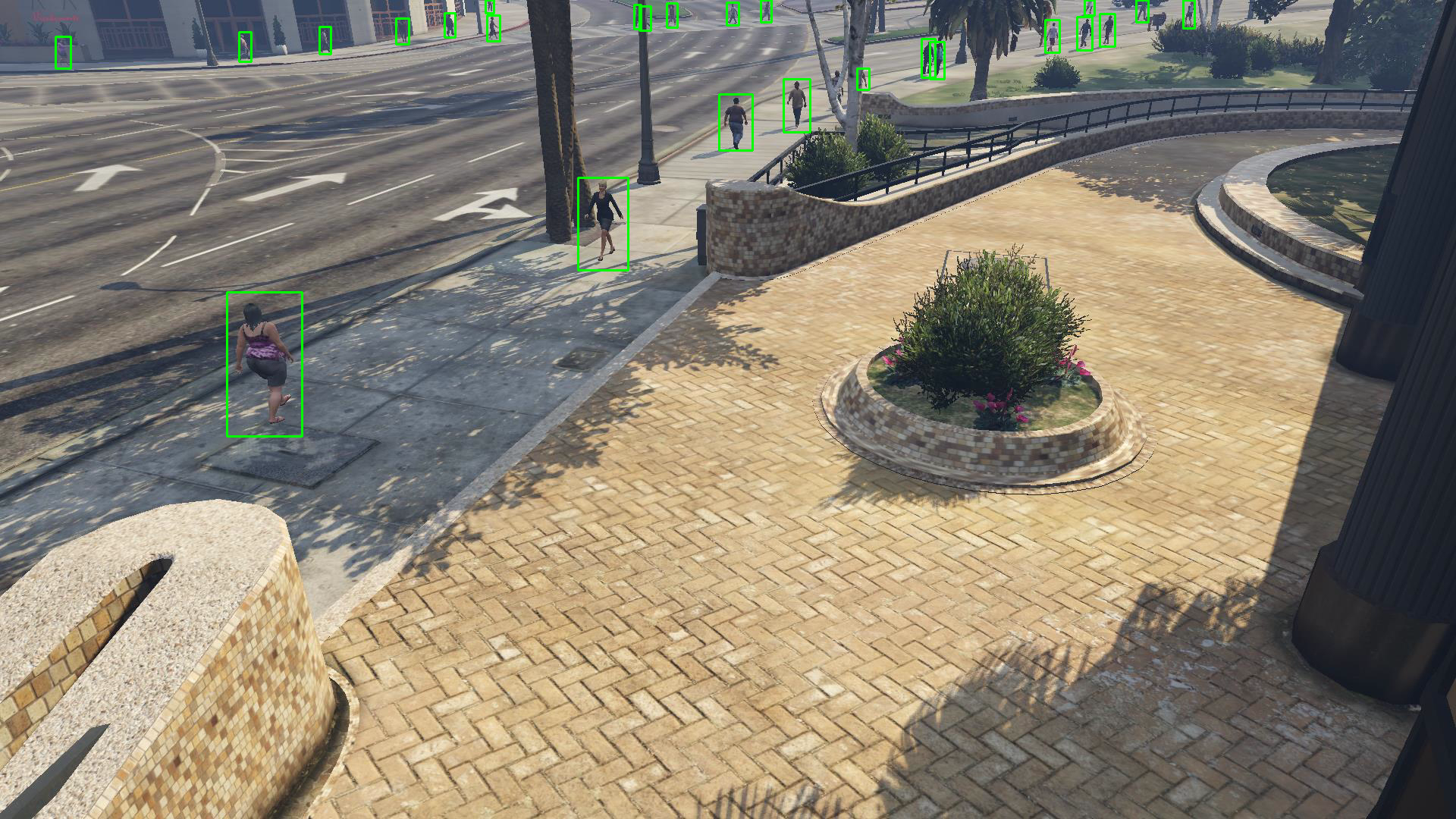}
        \caption{Normal: CAM 5}
    \end{subfigure}
    \begin{subfigure}[t]{0.32\columnwidth}
        \centering
        \includegraphics[width=\textwidth]{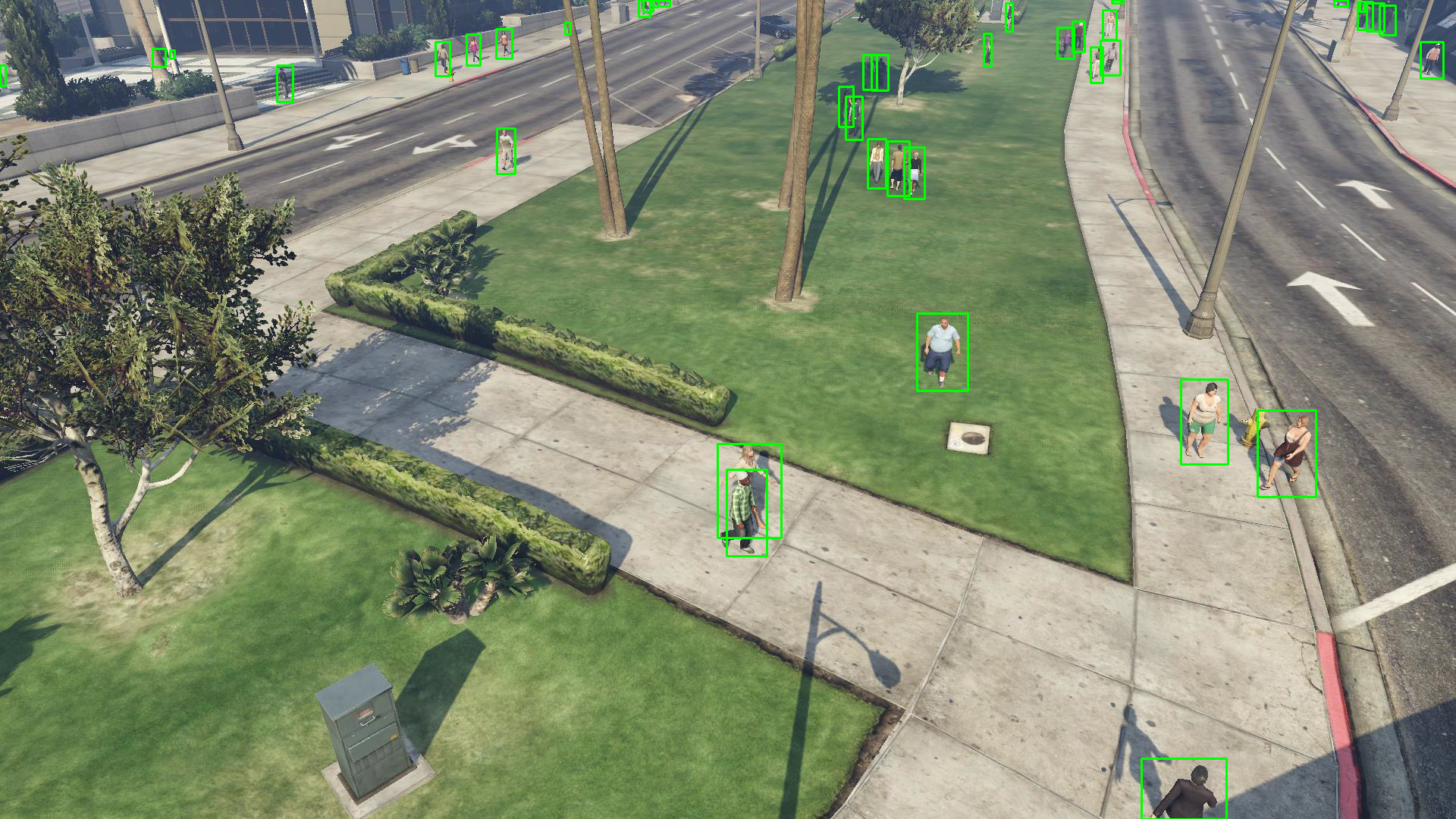}
        \caption{Hard: CAM 2}
    \end{subfigure}
    \caption{Example video images.}
    \label{fig:eval-video-info-samples}
\end{figure}

\begin{figure*}[t]
    \centering
    \includegraphics[width=1.9\columnwidth]{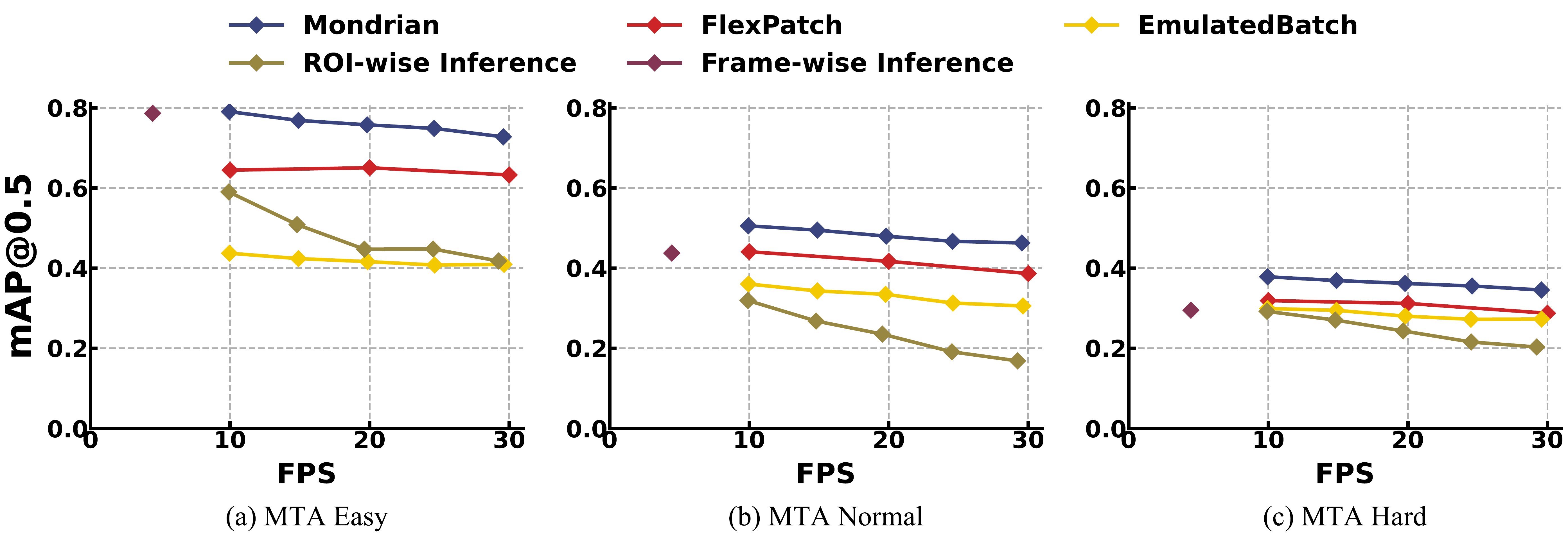}
    \caption{{\systemname} performance overview.}


    \label{fig:eval-tradeoff}
\end{figure*}

\subsection{Evaluation Setup}

\noindent{\bf Device.}
We implemented and evaluated {\systemname} on Samsung Galaxy S22 (Qualcomm Snapdragon 8 Gen 1 SoC, Adreno 730 GPU, Hexagon 780 DSP).
We use Android MediaCodec API for video decoding.

\vspace{3pt}
\noindent{\bf Object Detector.} We use multiple YOLOv5 object detectors with different sizes for evaluation (n < s < m < l < x).
We implement the detectors in TensorFlow 2.11.0 and convert them using TensorFlow-Lite 2.11.0~\cite{tflite}.
Unless stated otherwise, we report the results of our system using YOLOv5m with 640$\times$640 and 1280$\times$1280 input sizes over TensorFlow Lite on Samsung Galaxy S22. For multiple input sizes, Section \ref{subsubsec:scheduler-planning} describes the details.


\vspace{3pt}
\noindent {\bf Dataset.}
We use the MTA~\cite{mta} video dataset for evaluation.
MTA comprises 12 videos capturing people walking in the GTA simulator (each video has 4,922 frames, 1080p@30FPS).
There are 6 cameras and 2 videos for each camera.
As shown in Figure~\ref{fig:eval-video-info-stats}, we categorize the 6 cameras into 3 groups (MTA Easy, MTA Normal, MTA Hard) with difficulty.
Figure~\ref{fig:eval-video-info-samples} shows the example frames in different groups.
We use one video for training and one for testing each camera.
Furthermore, to reflect realistic deployment scenarios, we sample frames from the training dataset at one-minute intervals and train the scale estimator only with ten labeled frames for each camera.


\vspace{3pt}
\noindent {\bf Metric.}
We use two metrics to evaluate {\systemname}.

\begin{itemize}
    \item \textbf{Throughput (FPS)} is calculated as the total number of frames processed per second.
    \item \textbf{Accuracy (AP@0.5)} is evaluated using the average precision with IoU threshold 0.5 (AP@0.5) for each video and calculate the average across all videos.
\end{itemize}



\vspace{3pt}
\noindent{\bf Baseline.}
We compare {\systemname} with following schemes.

\begin{itemize}
    \item \textbf{FlexPatch~\cite{flexpatch}} is a detection-based tracking system that runs detection only for untrackable ROIs with packing.
    \item \textbf{Emulated Batching} is an ROI-based detection system that runs inference on multiple ROIs with emulated batching. Due to limited batch inference implementation on mobile deep learning frameworks, we emulated batching by dividing a large input tensor into grids. Specifically, we use 1280$\times$1280-sized input tensors containing 128$\times$128-sized ROIs inside (processing 100 ROIs in parallel).
    \item \textbf{ROI-wise Inference} extracts ROIs from the source Full HD input frames and runs the object detector for each ROI. We use a 128$\times$128 input size.
    \item \textbf{Frame-wise Inference} resizes the source Full HD input frames into the input size and runs the object detector on the entire frame. We use a 640$\times$640 input size, the default size of YOLOv5 models.
\end{itemize}

\subsection{Performance Overview}
\noindent{\bf End-to-End Throughput and Accuracy.}
Figure~\ref{fig:eval-tradeoff} shows the end-to-end performance of {\systemname} with YOLOv5m object detector over Samsung Galaxy S22 compared with baselines.
Overall, {\systemname} achieves the best mAP - throughput tradeoffs across all datasets and baselines.
(15.0-19.7\% higher mAP than FlexPatch, $\times$6.65 higher throughput than frame-wise inference).
The results well describe that {\systemname} is highly effective in curating the ROIs as reorganized inputs to maximize processing throughput.
Note that the mAP of Mondrian and FlexPatch are higher than the Frame-wise Inference in MTA Normal and MTA Hard because the detector errors are compensated by tracking.

\begin{table}[t]
\scriptsize
\centering
\captionsetup{skip=3pt}
\caption{Per-frame latency breakdown. The values denote the mean and standard deviation of the average latency per frame. Measured on the Galaxy S22 while processing CAM 3, the most complex video.}
\label{tb:eval-latency-breakdown}
\begin{adjustbox}{width=0.7\columnwidth}
\begin{threeparttable}
\begin{tabular}{ccc} 
    \toprule
    \textbf{System Component} & \textbf{Latency (ms)}     \\ \midrule
    PD ROI extraction         &  4.30 $\pm$ 2.66 \\ \midrule
    OF ROI extraction         &  3.77 $\pm$ 2.31 \\ \midrule
    ROI scale estimation      &  1.58 $\pm$ 1.05 \\ \midrule
    ROI postprocessing        &  4.06 $\pm$ 3.28 \\ \midrule 
    Canvas generation         &  8.65 $\pm$ 2.12 \\ \midrule
    *{\Packedinference}       & 27.24 $\pm$ 5.76 \\ \midrule
    Output reconstruction     &  0.91 $\pm$ 0.48 \\ \bottomrule
\end{tabular}
\begin{tablenotes}
  \tiny
  \item *~Packed inference runs on GPU, others on CPU.
\end{tablenotes}
\end{threeparttable}
\end{adjustbox}
\vspace{-5pt}
\end{table}

\vspace{3pt}
\noindent {\bf Per-frame Latency Breakdown.}
Table~\ref{tb:eval-latency-breakdown} shows the amortized per-frame latency breakdown of each step in the {\systemname} inference pipeline using YOLOv5m object detector with 640$\times$640 and 1280$\times$1280 input sizes on Samsung Galaxy S22.
Overall, we observe that the CPU components in the inference pipeline is well optimized, incurring minimal overhead.



\begin{figure}[t]
    \centering
    \includegraphics[width=1\columnwidth]{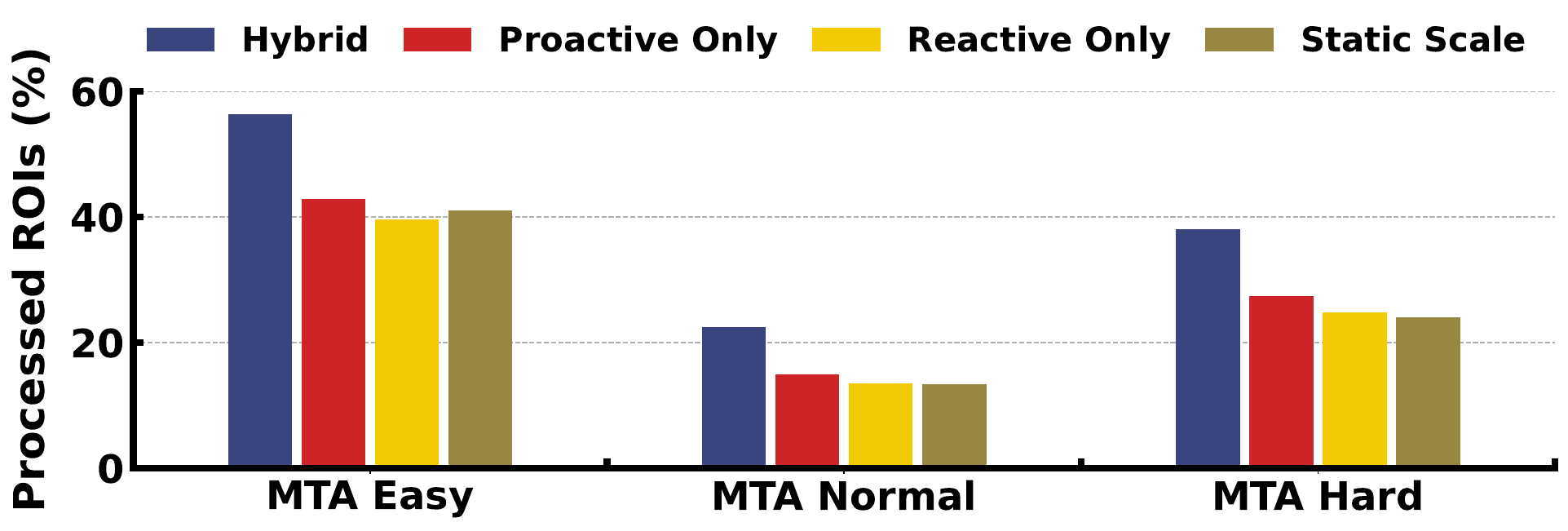}
    \caption{Performance of {\componentTwo}. Not-processed ROIs are dropped from packed canvases and interpolated after the inference.}
    \label{fig:eval-scaler}
\end{figure}

\subsection{Component-wise Evaluation}
\subsubsection{{\componentTwo}}
\label{subsubsec:eval-scale-estimator}

\begin{figure}[t]
    \centering
    \begin{subfigure}[t]{0.50\columnwidth}
        \centering
        \includegraphics[width=\columnwidth]{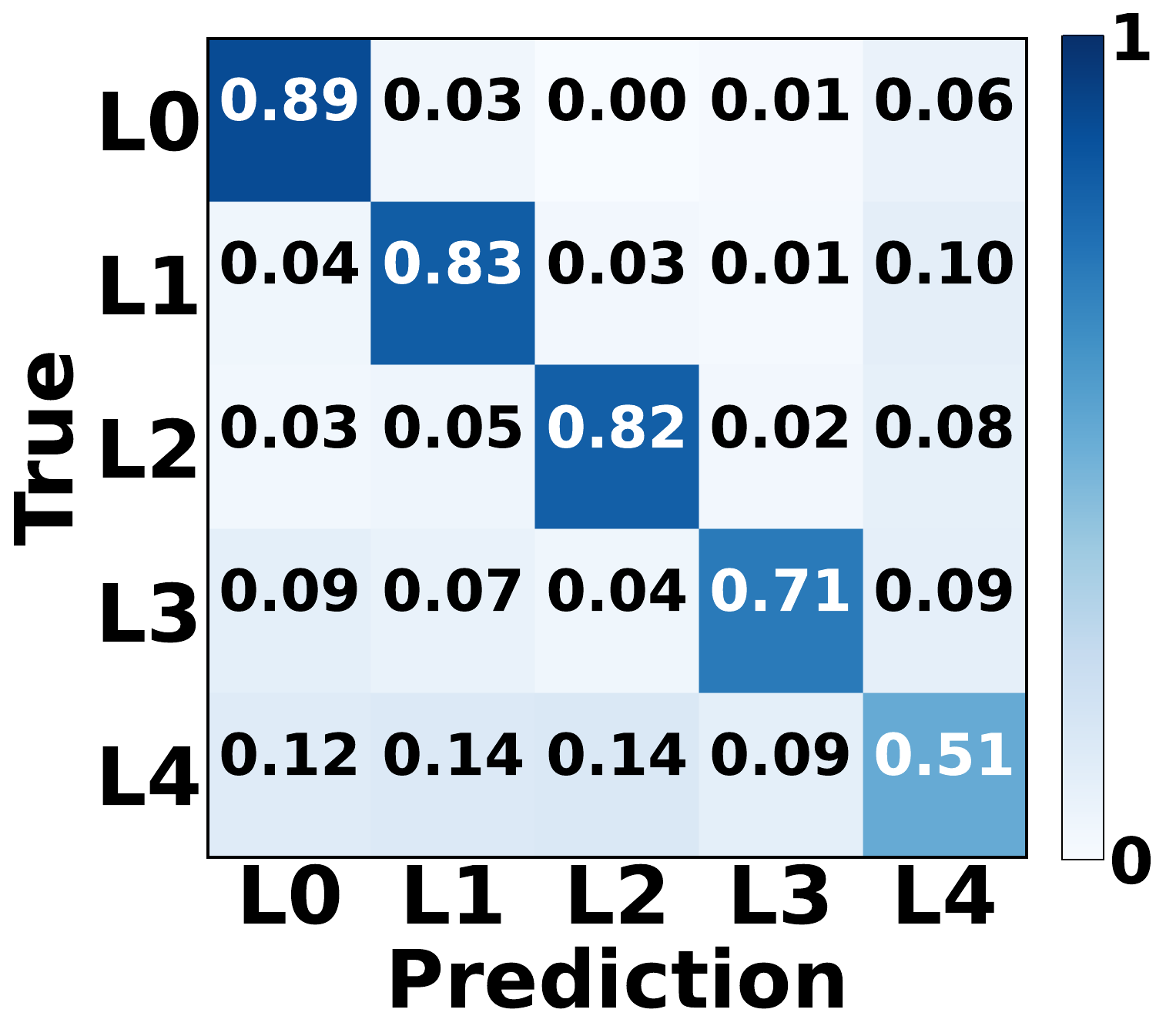}
        \caption{Confusion matrix}
        \label{fig:eval-scaler-proactive-cm}
    \end{subfigure}
    \begin{subfigure}[t]{0.48\columnwidth}
        \centering
        \includegraphics[width=1\columnwidth]{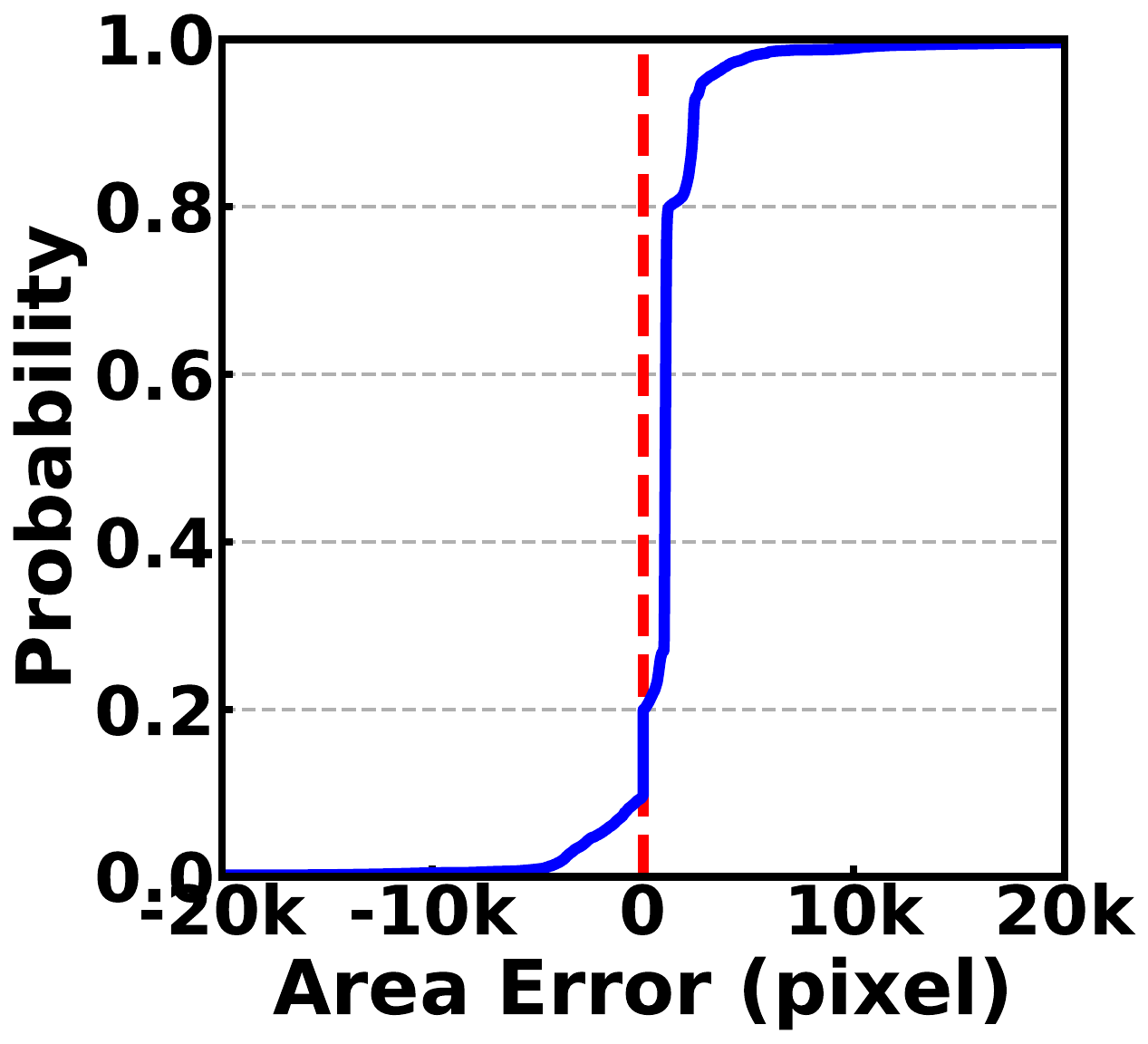}
        \caption{CDF of {\safearea} estimation error}
        \label{fig:eval-scaler-proactive-error-cdf}
    \end{subfigure}
    \caption{Accuracy of proactive scale predictor. {\Safearea} estimation error means the difference between the predicted and the ground truth {\safearea}.}
    \label{fig:eval-scaler-proactive}
\end{figure}

\noindent{\bf Effectiveness of Hybrid Approach.} 
Figure~\ref{fig:eval-scaler} shows the {\componentTwo}'s performance.
We compare it against three baselines: \textit{Proactive Only}, which uses the scale estimated by the ML predictor without further probing; \textit{Reactive Only}, which probes five scale candidates (0.5 to 0.9 with 0.1 intervals); and \textit{Static Scale}, which always scales ROIs into the 10000-pixel area.
The maximum ratio of processed ROIs in our approach (\textit{Hybrid}) shows that our {\componentTwo} can reduce the workload most effectively.
\textit{Reactive Only} suffers from a significant throughput drop due to probing overhead, while \textit{Prediction Only} can only partially utilize the scaling opportunity due to inaccurate prediction of the {\safearea}.
On the contrary, our hybrid scale estimation algorithm achieves high throughput and accuracy by fusing both approaches to search the {\safearea} with minimal overhead.
Note that MTA Hard shows a higher processed ROI ratio even with the largest number of ROIs due to ROIs' small size.

\noindent {\bf Proactive Scale Prediction Accuracy.} 
Figure~\ref{fig:eval-scaler-proactive} shows the performance of the proactive ROI scale predictor.
Specifically, Figure~\ref{fig:eval-scaler-proactive-cm} shows the confusion matrix of the proactive scale predictor on the MTA dataset, and Figure~\ref{fig:eval-scaler-proactive-error-cdf} shows the CDF of the prediction error.
They show that the prediction is highly accurate, with 90\% of ROIs showing errors of less than 10 pixels.
More importantly, underestimation cases (i.e., predicting the scale smaller than the ground truth, leading to object detection failure) rarely occur (less than 10\%).
Note that the reactive scale tuner will further mitigate the overestimation cases.

\begin{figure}[t]
    \centering
    \includegraphics[width=0.99\columnwidth]{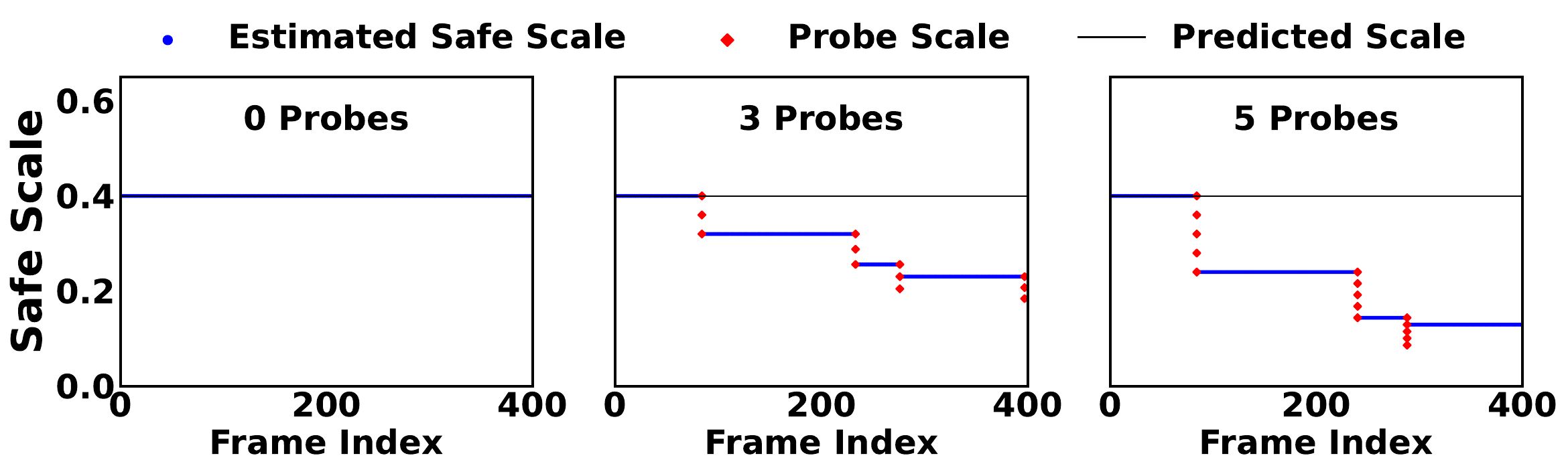}
    \caption{Reactive scale tuner example operation.}
    \label{fig:eval-scaler-reactive-timeline}
\end{figure}

\noindent {\bf Reactive Scale Tuning Overhead.}
As reactive tuning runs once every schedule, the number of ROIs for probing is much smaller than the total number of ROIs.
Moreover, while many candidates incur considerable overhead in the first window, the overhead quickly reduces in the subsequent windows as the optimal scale is more likely to be searched.
Figure~\ref{fig:eval-scaler-reactive-timeline} shows the {\estimatorTwo}'s operation timeline.
Starting from the estimated {\safearea} from the proactive predictor, it successfully probes the neighboring candidates to optimize the {\safearea} further and improve throughput.

\begin{figure}[t]
    \centering
    \begin{subfigure}[t]{0.54\columnwidth}
        \centering
        \includegraphics[width=\columnwidth]{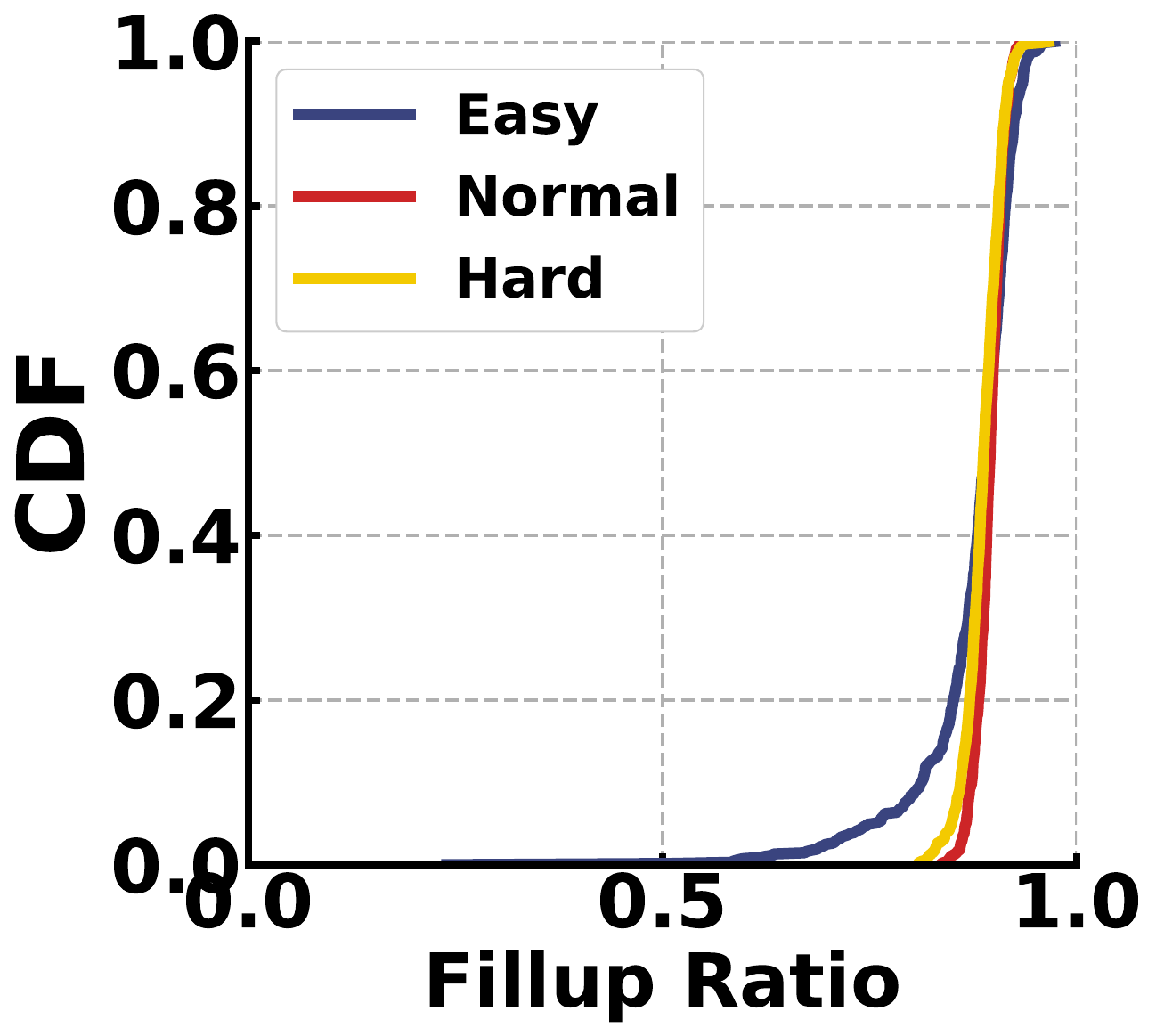}
        \caption{CDF of canvas fillup ratio}
        \label{fig:eval-scheduler-fillup-ratio}
    \end{subfigure}
    \begin{subfigure}[t]{0.42\columnwidth}
        \centering
        \includegraphics[width=\columnwidth]{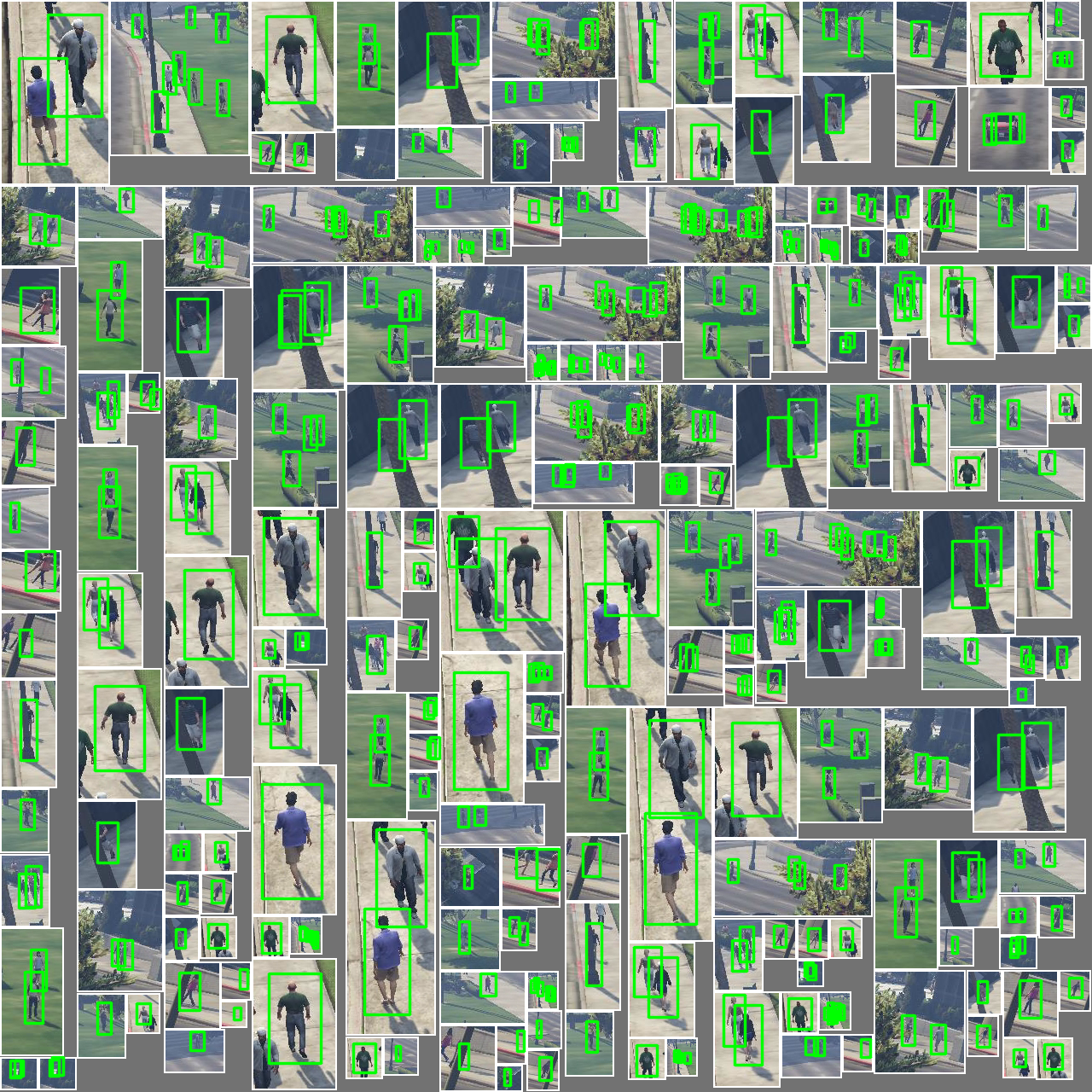}
        \caption{Example packed canvas (Fill up ratio: 86.8\%)}
        \label{fig:eval-scheduler-packed-canvas-example}
    \end{subfigure}
    \caption{Performance of {\componentThree}.}
    \label{fig:eval-scheduler}
\end{figure}

\begin{figure}[t]
    \centering
    \includegraphics[width=1\columnwidth]{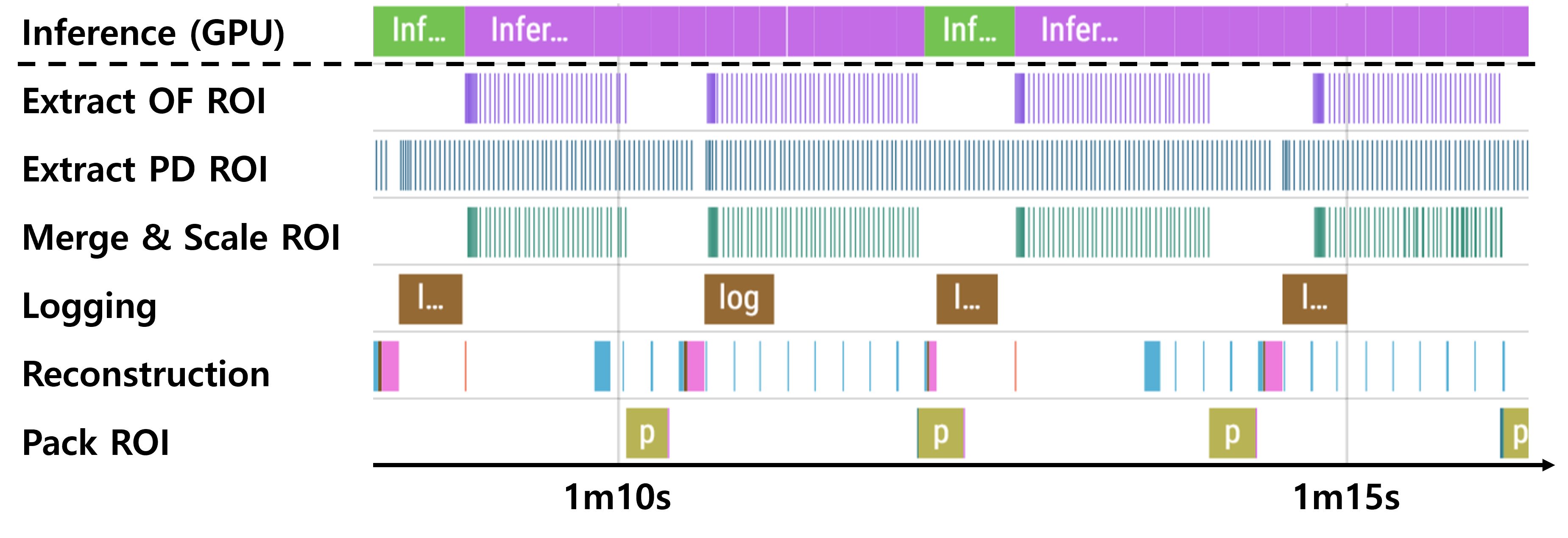}
	\caption{Scheduling timeline of the {\systemname} system. Only Inference runs on GPU, and others run on CPU. For Inference, green box mean inference on a full frame, pink box means inference on a packed canvas.}
    \label{fig:timeline}
\end{figure}

\subsubsection{\componentThree}

Figure~\ref{fig:eval-scheduler} shows how the {\componentThree} efficiently packs the ROIs to maximize the pixel processing throughput.
{\systemname} tightly packs the ROIs into the canvases with mostly 80-90\% fillup ratio.
Figure~\ref{fig:eval-scheduler-packed-canvas-example} shows that the 86.8\% filled canvas is highly dense.
Moreover, a pipelined system structure hides the packing time (Figure~\ref{fig:timeline}).


\begin{table}[h]
\small
\centering
\captionsetup{skip=3pt}
\caption{Maximum memory usage and CPU utilization.}
\label{tb:eval-cpu-mem-overhead}
\begin{adjustbox}{width=0.95\columnwidth}
\begin{tabular}{cc|cccc} 
\toprule
\singlecol{2}{Input\\Resolution} & \singlecol{2}{Schedule\\Interval} & \singlecol{2}{Total\\Memory} & \singlecol{2}{Detector\\Memory} & \singlecol{2}{Image\\Memory} & \singlecol{2}{CPU} \\
                                 &                                   &                              &                                 &                              &                    \\ \midrule
640$\times$640                   & 2s                                & 2.5 GB                       & 1.5 GB                          & 1.0GB                        & 32\%               \\ \midrule
640$\times$640                   & 4s                                & 2.9 GB                       & 1.5 GB                          & 1.3GB                        & 31\%               \\ \midrule
1280$\times$1280                 & 2s                                & 2.9 GB                       & 1.8 GB                          & 1.1GB                        & 32\%               \\ \midrule
1280$\times$1280                 & 4s                                & 3.2 GB                       & 1.8 GB                          & 1.4GB                        & 33\%               \\ \bottomrule
\end{tabular}
\end{adjustbox}
\vspace{-10pt}
\end{table}

\subsection{Resource Overhead}

\noindent{\bf Memory Usage and CPU Utilization.} 
Table~\ref{tb:eval-cpu-mem-overhead} shows {\systemname}'s resource overhead regarding memory usage and CPU utilization.
Overall, the overhead ($\approx$30\% CPU utilization and $\approx$3~GB memory) is relatively low, indicating that {\systemname} can be effectively deployed on resource-constrained edge devices.
The memory usage is mainly from keeping the multiple streams of high FPS videos (e.g., storing 30 frames of a 1080p video requires $\approx$180~MB) and the object detectors.
If the system does not have to generate output video, we can discard the images right after packed canvas generation, leading to less memory consumption from images.



\section{Related Works}

\noindent {\bf Live Video Analytics.} Live video analytics applications are continuously emerging in various application domains such as smart homes, elderly care, surveillance, and transportation \cite{zhang2019edge, eldib2016behavior, prati2019sensors, balasundaram2020intelligent, gautam2019video, ananthanarayanan2017real, barthelemy2019edge, naphade20215th, chang2020ai}. 
In this work, we aim to address challenges in relatively under-explored scenarios where high-performance video analysis over high-resolution video input should be enabled only with the local processing capabilities of resource-limited edge devices (as indicated in Section~\ref{subsec:motiv-scenarios}). 

\vspace{3pt}
\noindent {\bf Lightweight Object Detection Models.} 
Many works aim at improving the computation efficiency (in terms of FLOPs or weights) of the object detectors. Single-pass detectors such as SSD~\cite{liu2016ssd} or YOLO~\cite{yolov4} remove inefficiencies in prior two-pass detectors such as FasterRCNN~\cite{ren2015fasterrcnn}. 
Recent works like EfficientDet~\cite{tan2020efficientdet} utilize state-of-the-art lightweight convolutional neural networks as their backbone (e.g., EfficientNet~\cite{tan2019efficientnet} applying the compound scaling of layer depth, channel width, and input resolution). 
We take an orthogonal approach to further accelerate object detection speed, which is applicable regardless of underlying detectors. 

\vspace{3pt}
\noindent {\bf ROI-based Processing.} 
There have been several techniques for efficient pre-processing of input data including, background removal with edge detector~\cite{yi2020eagleeye}, degrading compression quality for non-ROI areas~\cite{uzkent2020learning}, scaling input images~\cite{wang2019elastic, zhu2021dynamic}, packing with extracted ROIs~\cite{gokarn2023mosaic}. Also, a work applies different DNNs for ROIs with different complexity~\cite{remix}. Our work is complementary to such approaches; we dynamically resize input ROIs considering dynamic content changes and merge them to further improve processing efficiency.

\vspace{3pt}
\noindent {\bf Acceleration with Object Tracking.} A set of work accelerates the object detection task over streaming video data by lightweight tracking across consecutive video frames ~\cite{apicharttrisorn2019frugal, liu2020continuous, he2021real, flexpatch}.
Different from these works, we take a frame-by-frame processing approach to achieve higher detection accuracy even under fast-changing scenes with many small objects where tracking suffers from severe accuracy drop. It instead overcomes the performance burden of frame-by-frame processing by carefully-designed ROI rescaling and packing. 

\vspace{3pt}
\noindent {\bf Acceleration via Computation Offloading.} Another widely adopted acceleration approach is offloading computation-intensive detection tasks to the cloud or edge servers. The concept of ROI has been widely adopted in these systems to transfer less amount of data and thus achieve a smaller end-to-end latency ~\cite{zhang2021elf, liu2019edge,du2020server}. 
Another thread of works is proposed for efficient collaboration between camera devices and servers to achieve high processing throughput ~\cite{kang2017neurosurgeon, laskaridis2020spinn, almeida2021dyno}. 
Our system is designed with a different goal, i.e., enabling high-performance analysis only using local computing capability, and can better suit the scenarios requiring low operational cost and having high privacy concerns. 


\vspace{3pt}
\noindent {\bf Multi Object Tracking.} 
Several frameworks are proposed to address the Multi-object tracking (MOT) problem~\cite{wang2020towards, dendorfer2021motchallenge}, 
MOT consists of two stages: detection and re-identification. The re-identification is often heavier than detection to identify objects with limited information, making it challenging to apply on resource-constrained devices. Even lightweight MOT systems~\cite{dong2021polarmask, shuai2021siammot, tsai2022mobilenet} require high-capacity devices. We focus on edge devices with restricted resources and enables frame-by-frame detection in soft real-time, which can further be sued for optimizing MOT on edge devices.

\section{Discussion and Future Works}

\noindent {\bf Incorporation of Other Accelerators.} We are extending {\systemname} to support various processors such as Neural Processing Units (NPUs) or Tensor Processing Units (TPUs). They also have similar characteristics to GPUs where their utilization becomes much higher with sufficient input workload. 
In addition, we can easily port and deploy our system to more powerful edge devices such as Jetson TX2 or AGX Xavier.
We expect to achieve a similar level of performance gain 
while we could further optimize our techniques to utilize the parallel processing capabilities (e.g., Nvidia Hyper-Q~\cite{HyperQ}) supported by more advanced hardware. Such generalizability toward various processors is useful as the multiple hardware architectures still compete for edge devices. 

\vspace{3pt}
\noindent {\bf Using ROI Scale Estimator for Offloading.} Our component techniques can be leveraged in other systems. For instance, the ROI extractor and scale estimator can be applied to reduce data size to transfer in cloud offloading systems. 
Some works extract ROIs on the edge devices and only sends ROIs to the cloud~\cite{liu2019edge, yi2020eagleeye, liu2016ssd}. 
ROI scale estimator can provide a new capability to dynamically reduce the ROI sizes based on the input scenes, which widens the benefits of cloud offloading. More detailed studies remain as our future work. 

\vspace{3pt}
\noindent {\bf Scheduling Policy and Algorithm.} We showed that ROI packing is an NP-hard problem and suggested a greedy approximation solution. Our framework is designed to facilitate the implementation of various scheduling policies and algorithms, which we plan to study in future work.

\section{Conclusion}

We presented {\systemname}, an edge system that enables high-performance object detection on high-resolution video streams.
To enable such capability, we devise a novel \textit{{\approachname}} that significantly enhances throughput by processing the smallest possible combinations of input pixels.
In particular, we devised two core techniques: (i) {\componentTwo} to dynamically reduce the ROIs to the \emph{\safeareas} without compromising detection accuracy, (ii) throughput-maximizing {\componentThree} to determine the optimal combination of canvases and pack rescaled ROIs into them.
{\systemname} outperforms the state-of-the-art techniques (e.g., input rescaling, ROI extractions, ROI extractions+batching) by 15.0-19.7\% higher accuracy, leading to $\times$6.65 higher throughput than frame-wise inference for processing various 1080p video streams.

\section*{Acknowledgement}
This work was supported by the National Research Foundation of Korea (NRF) grant funded by the Korea government (MIST) (No. 2022R1A2C3008495).

\bibliographystyle{IEEEtran}
\bibliography{references}

\begin{IEEEbiography}
[{\includegraphics[width=1in,height=1.25in,clip,keepaspectratio]{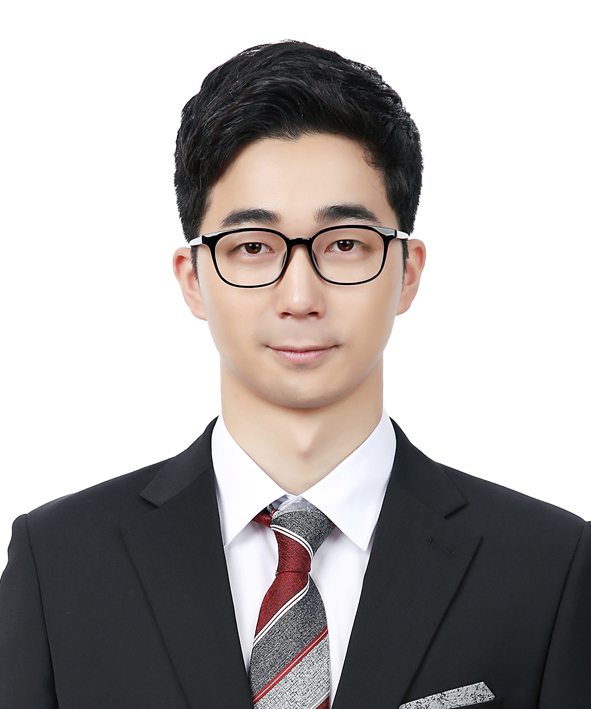}}]
{Changmin Jeon} is a Ph.D. student at the Department of Computer Science and Engineering, Seoul National University, Korea. He received his B.S degree in Mechanical and Aerospace Engineering and Computer Science and Engineering from Seoul National University in 2020. His research interests include mobile/embedded deep learning systems and eXtended Reality (XR).
\end{IEEEbiography}

\begin{IEEEbiography}
[{\includegraphics[width=1in,height=1.25in,clip,keepaspectratio]{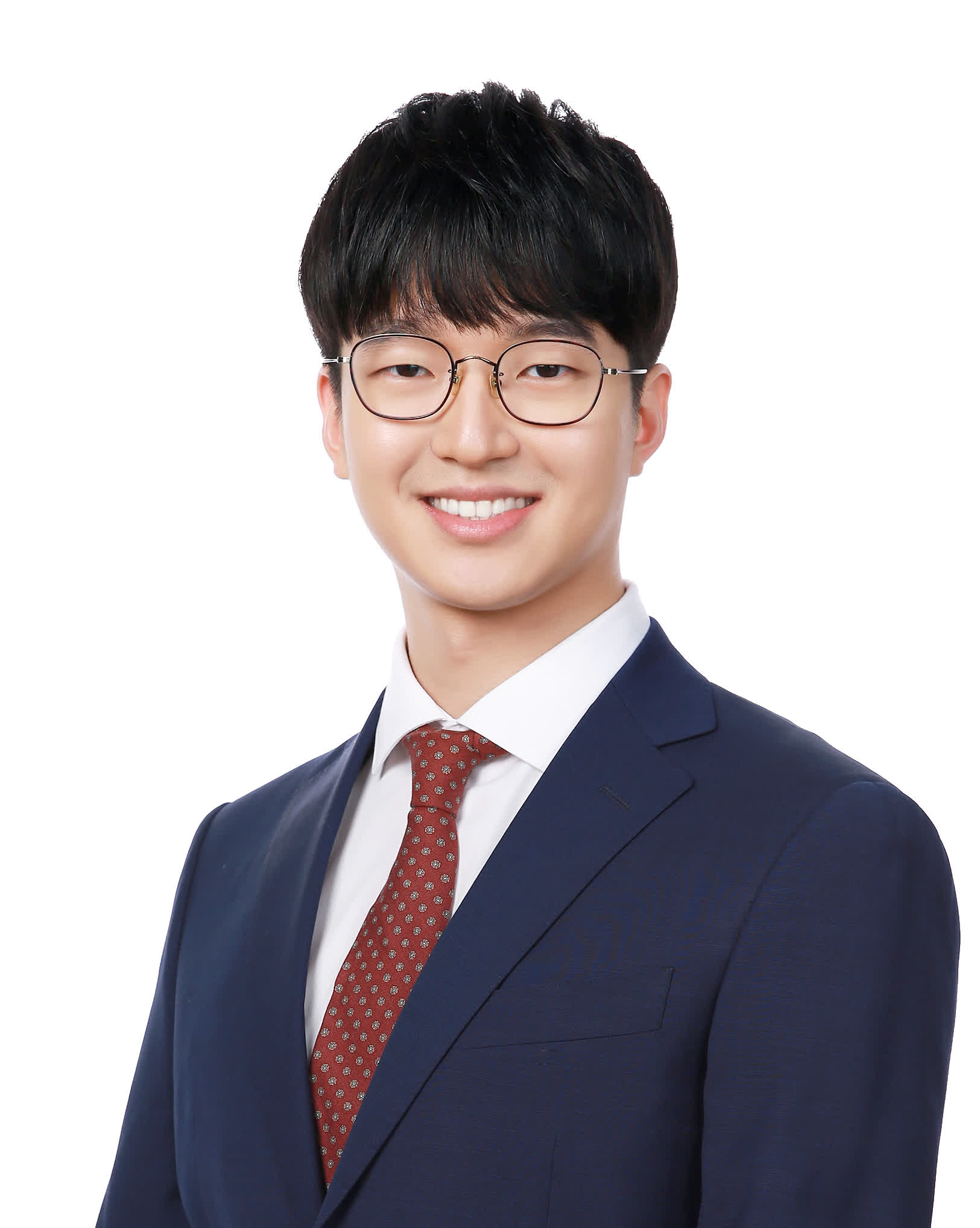}}]
{Seonjun Kim} is a Ph.D. student at the Department of Computer Science and Engineering, Seoul National University, Korea. He received his B.S. in Electrical and Computer Engineering from Seoul National University in 2022. His research interests lies in algorithm-system co-optimization for eXtended Reality(XR) applications on mobile devices.
\end{IEEEbiography}

\begin{IEEEbiography}
[{\includegraphics[width=1in,height=1.25in,clip,keepaspectratio]{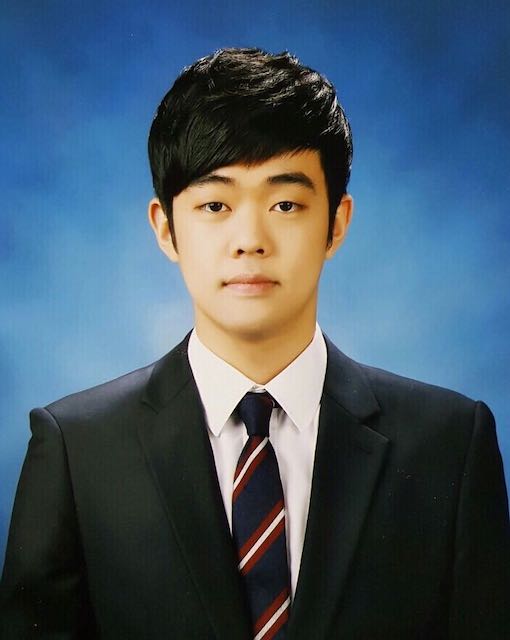}}]
{Juheon Yi} is a Research Scientist at Nokia Bell Labs, Cambridge, UK. He received his Ph.D. in Computer Science and Engineering from Seoul National University in 2024. His research interests include mobile/embedded AI and video analytics systems. He is a recipient of the Microsoft Research Asia Ph.D. Fellowship 2020 and Best Paper Award at the Students in MobiSys 2021 Workshop.
\end{IEEEbiography}

\begin{IEEEbiography}
[{\includegraphics[width=1in,height=1.25in,clip,keepaspectratio]{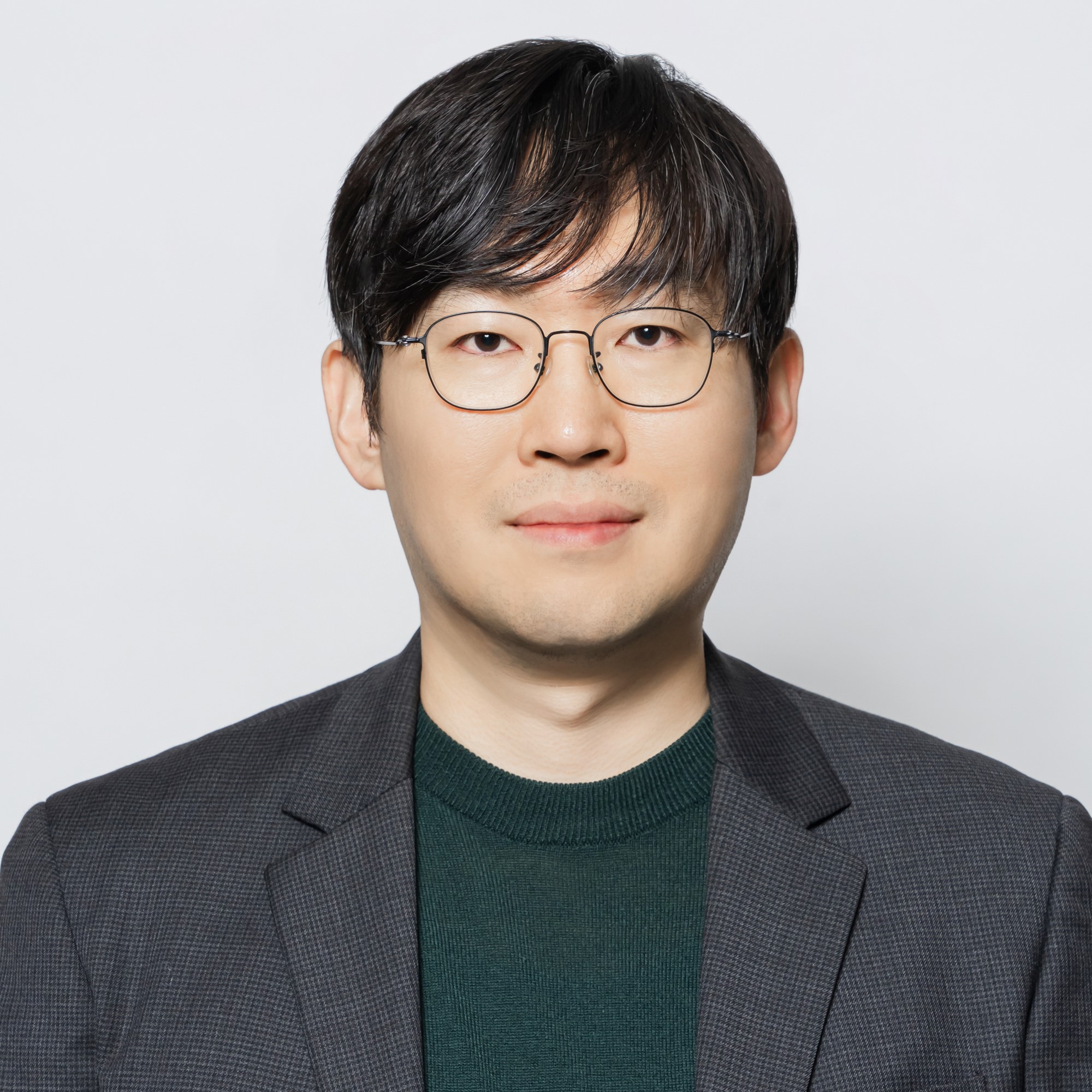}}]
{Youngki Lee} is an associate professor at Seoul National University. His research interests include innovative software systems over a wide spectrum across systems, applications, and users. He received a Ph.D. degree in computer science from KAIST in 2013. Contact him at youngkilee@snu.ac.kr.
\end{IEEEbiography}

\end{document}